\documentclass[10pt,twocolumn,letterpaper]{article}

\usepackage[pagenumbers]{wacv}

\usepackage{multirow}
\usepackage{pifont}
\usepackage{pifont}
\usepackage{booktabs}
\usepackage[table]{xcolor}
\usepackage{array}
\usepackage{makecell}
\usepackage{wrapfig}

\newcommand{\sectionrow}[2]{%
  \arrayrulecolor{black}%
  \rowcolor{secbar}\multicolumn{#1}{c}{\textit{#2}} \\
}

\definecolor{ourshl}{HTML}{E8F1FA}
\definecolor{secbar}{HTML}{DDDDDD}
\definecolor{medbg}{HTML}{FFEEF4}
\colorlet{medbglight}{medbg!50}
\definecolor{natbg}{RGB}{220,240,255}

\usepackage{pgfplots}
\pgfplotsset{compat=1.18}
\usetikzlibrary{arrows.meta}

\usetikzlibrary{calc}

\definecolor{wacvblue}{rgb}{0.21,0.49,0.74}
\usepackage[pagebackref,breaklinks,colorlinks,allcolors=wacvblue,bookmarks=False]{hyperref}

\def\confName{WACV}
\def\confYear{2027}

\title{
From Natural Images to Medical Scans: \\
Transferability of Diffusion Models for Medical Image Generation
}

\author{
Efthymios Georgiou$^{1,2}$ \quad
Ioannis Kakogeorgiou$^{3}$ \quad
Sarah Br\"uningk$^{1,2}$\\[2pt]
{\small $^1$Department of Digital Medicine, University of Bern, Switzerland} \\
{\small $^2$Department of Radiation Oncology, Inselspital, Bern University Hospital and University of Bern, Switzerland}\\
{\small $^3$IIT, National Centre for Scientific Research “Demokritos”, Greece}
}

\newcommand{\ttt}[1]{\texttt{#1}}

\newcommand{\alert}[1]{{\color{red}{#1}}}

\newcommand{\Th}[1]{\textsc{#1}}

\newcommand{\ch}{\checkmark}
\newcommand{\xmark}{\ding{55}}

\newcommand{\citeme}[1]{\alert{[X]}}
\newcommand{\refme}[1]{\alert{(X)}}

\makeatletter
\newcommand*\bdot{\mathpalette\bdot@{.7}}
\newcommand*\bdot@[2]{\mathbin{\vcenter{\hbox{\scalebox{#2}{$\m@th#1\bullet$}}}}}
\makeatother

\makeatletter
\DeclareRobustCommand\onedot{\futurelet\@let@token\@onedot}
\def\@onedot{\ifx\@let@token.\else.\null\fi\xspace}

\def\eg{\emph{e.g}\onedot} 
\def\ie{\emph{i.e}\onedot}

\def\etal{\emph{et al}\onedot}
\makeatother

\definecolor{ForestGreen}{RGB}{34,139,34}

\definecolor{DarkenedMagenta}{RGB}{225,0,167}

\newcommand{\LDSc}{\Th{LDS\footnotesize{c}}\xspace}

\begin{document}
\maketitle

\begin{abstract}
Transfer learning from natural images is standard in supervised and increasingly adopted in self-supervised medical image analysis, but its role in generative modelling, and in particular diffusion denoisers, remains underexplored. Medical diffusion models typically train the denoiser from scratch, with no systematic evaluation of natural-image transfer. We analyse this across latent and pixel paradigms, model sizes (131M--952M) and two radiology datasets (RadImageNet, BraTS), and ask which aspects of transfer support generation. We find that: i) Natural-image initialisation consistently improves generation, by 21-34\% relative gFID in pixel space against 14\% in latent space, where the tokeniser supplies much of the prior. gFID gains mirror downstream pathology classification and radiomic-feature FD, both proxies for clinical fidelity. ii) Benefits reflect spatial structure, not semantics: semantic probing does not predict generation quality, whereas
a spatial-structure metric adapted to internal denoiser representations tracks gFID at $|r|{=}$0.85, positioning spatial organisation as a channel through which pretraining manifests. iii) Mixed-domain denoiser training shows that natural and medical images share similar layer-wise representations up to the late blocks, and mixing natural images into medical training improves fidelity. iv) Transfer benefit is tied to natural image initialisation, outperforming in-domain medical pretraining and several reinitialisation schemes.
Overall, natural image denoisers transfer positively to medical ones in our setup. Code is available  at \href{https://github.com/CAIROLab-Bern/rad-dit}{CAIROLab-Bern/rad-dit}.
\end{abstract}

\section{Introduction}
\label{sec:intro}

Pedagogical drawing textbooks decompose image creation into a small set of basic primitives, \eg, edges, contours, proportions, and light-and-shade, from which complex scenes can be composed ~\cite{nicolaides1941natural,edwards1979drawing}. A related principle appears in visual neuroscience, where neurons in early visual cortex respond to low-level features such as oriented edges~\cite{hubel1962receptive,kamitani2005decoding}, which are present in any visual input, whether a natural photograph or a medical scan.

\emph{Domain expertise is built on top of this shared perceptual basis.}
Radiologists, for example, acquire domain-specific pattern-recognition skills through extensive exposure to medical images~\cite{waite2019perceptual,alexander2020radiologists}, but this advantage does not generally extend to non-radiological visual search tasks~\cite{nodine1998perceptual}. This suggests a distinction between broadly reusable visual primitives and learned domain-specific expertise.

A similar pattern motivates transfer learning in computer vision.
Initialising from \Th{ImageNet} (IN)~\cite{deng2009imagenet} is the standard approach in \emph{supervised} medical imaging. Early works on convolutional models find that transfer benefits are modest, mostly faster convergence and reuse of low-level (generic) features~\cite{raghu2019transfusion,ke2021chextransfer}. Mustafa~\etal~\cite{mustafa2021supervised} identify scale as a crucial factor for transferability. Matsoukas~\etal~\cite{matsoukas2022makes} extend
their analysis to convolutional~\cite{fukushima1980neocognitron, lecun1989backpropagation} and transformer~\cite{vaswani2017attention, dosovitskiy2021an} models and find feature reuse to be the main source when transfer helps, especially for smaller datasets and closer domains. Interestingly, transformers rely on feature reuse more than convolutional networks.
In \emph{self-supervised} (SSL) medical imaging, natural-image encoders are increasingly used as initialisation or off-the-shelf, at times matching domain-specific models~\cite{wu2026braindino, karasikov2025midnight}.
For \emph{generative} modelling, however, and diffusion denoisers in particular, the role of transfer remains underexplored.

State-of-the-art diffusion generative models fall into two broad paradigms: \emph{pixel diffusion} and \emph{latent diffusion}.
\emph{Pixel diffusion} applies the denoiser directly to pixels.
\emph{Latent diffusion} pairs a tokeniser with a denoiser that operates in its latent space.
In \emph{latent diffusion}, natural-image priors can be transferred through the tokeniser, the denoiser, or both.
In the medical domain, natural-image tokenisers have been reused, and a few works fine-tune pretrained text-to-image denoisers~\cite{chambon2022roentgen,yellapragada2024pathldm}.
Yet denoiser transfer remains uncommon~\cite{pinaya2022brain, huang2024chest, guo2025maisi, liu2025treatmentaware, wang20253d} and is treated as an implementation detail rather than a studied phenomenon.
Whether natural-image denoiser \emph{transfer is beneficial} for medical image generation, and \emph{which aspects} of the pretrained model account for any observed benefit, remain open questions.

In this work, we study natural-image denoiser transfer for medical image
generation. Our analysis revolves around two broad questions: do natural
denoisers transfer to the medical domain, and if so,
what is it that transfers and how?
We address these across latent and
pixel-space, model sizes ($131$M--$952$M), and two radiology
datasets (\Th{RadIN}; \Th{BraTS}).
Our results show a positive transfer, and suggest that the benefit is tied to natural-image initialisation and manifests as better spatial organisation of the internal denoiser representations. We centre our analysis on pixel denoisers, with parallel results for latent ones.

\noindent Our main findings are:
\begin{enumerate}
  \item Natural-image initialisation improves generation in both paradigms. Pixel-space models benefit most, improving by
  $21$--$34\%$ gFID against $14\%$ in latent space, where the tokeniser carries much of the prior. FID gains are mirrored in radiomic Fréchet distance (\Th{FD-Rad}) and pathology classification \Th{macroF1}, both proxies for clinical fidelity.

  \item Transfer benefits reflect spatial structure. Semantic probing shows that linear separability does not predict generation quality, while a modified spatial structure metric for diffusion representations, \LDSc, tracks gFID at $|r|{=}0.85$, positioning spatial organisation as a channel through which pretraining manifests.

  \item Mixed-domain experiments show that the denoiser computes similar layer-wise representations for natural and medical inputs through most of the network, and mixing natural data into medical training improves medical fidelity.

  \item The benefit is tied to natural-image initialisation. It cannot be recovered by medical pretraining or several re-initialisation schemes. Illustrative extrapolation suggests far more medical data would be needed to close the gap.

\end{enumerate}

\section{Related Work}

\subsection{Diffusion and Flow Models}
Diffusion~\cite{sohl2015deep, ho2020denoising, song2021scorebased} and flow models~\cite{liu2023flow, lipman2023flow} generate data from noise.
To do so, they are trained to reverse a forward noising
process of data towards noise. The networks that approximate this denoising process are \emph{denoisers}.

\noindent
\textbf{Latent-space denoising.}
Latent denoisers operate in the latent space of a visual tokeniser, typically a Variational Autoencoder
(\ttt{VAE})~\cite{kingma2014vae}, which reduces spatial
dimensionality and computational cost relative to pixel-space diffusion.
Esser~\etal~\cite{esser2021taming} propose a \ttt{VQ-GAN} to compress the image space into a discrete codebook, trained with additional perceptual and patch discriminator~\cite{goodfello2014gans} loss.
Rombach~\etal~\cite{rombach2022high} use a continuous, KL-regularised \ttt{VAE} latent space with a \ttt{U-Net} denoiser~\cite{ronneberger2015unet}.
\ttt{VA-VAE}~\cite{yao2025reconstruction} improves the latent space structure while achieving larger spatial compression (16$\times$), by aligning it with \ttt{DINOv2} features~\cite{oquab2024dinov2}.
On the denoiser side, Peebles~\etal~\cite{peebles2023scalable} replace the
\ttt{U-Net} backbone with a Diffusion Transformer (\ttt{DiT}), and \ttt{SiT}~\cite{ma2024sit}
extends it to flow objectives. \ttt{LightningDiT} (\ttt{LDiT})~\cite{yao2025reconstruction} proposes training and architectural advancements on top of the vanilla \ttt{DiT} architecture.
Here, we couple \ttt{LDiT} with different \ttt{VAE}s, and study whether the denoiser benefits from natural-image transfer beyond the prior already provided by the tokeniser.

\noindent
\textbf{Pixel-space denoising.}
Pixel diffusion applies the denoiser directly to pixels, avoiding the
\ttt{VAE} reconstruction (decoder) bottleneck. Early diffusion models operate in
pixel space~\cite{ho2020denoising}, but the high dimensionality led to the popularisation of the latent paradigm~\cite{esser2021taming, rombach2022high}. Pixel-based approaches
have also been used as multi-stage cascades for increasing resolution~\cite{ho2022cascaded, saharia2022photorealistic}.
Only recently, \ttt{JiT}~\cite{li2026jit} revived single-stage pixel diffusion with a plain transformer that
denoises large pixel patches (16--32), carrying over advances from \ttt{DiT} and an
$\mathbf{x}$-prediction objective.
We use \ttt{JiT} as our pixel-space denoiser, and, to the best of our knowledge, are the first to transfer it to the medical domain.

\noindent
\textbf{Representation Alignment \& Joint denoising.}
\ttt{REPA}~\cite{yu2025representation} aligns
diffusion features to a pretrained encoder, and \ttt{iREPA}~\cite{singh2026what} finds that spatial structure, rather than semantics, drives alignment gains.
We observe an analogous pattern for natural-image initialisation.
Joint-denoising methods model images and features together.
\ttt{ReDi}~\cite{kouzelis2025redi} couples \ttt{VAE} latents with PCA \ttt{DINOv2} features, and \ttt{REGLUE}~\cite{petsangourakis2026reglue} denoises compressed \ttt{DINOv2} features concatenated with \ttt{VAE} latents.
\ttt{Latent Forcing}~\cite{baade2026latent} brings joint denoising to pixel space, and \ttt{V-Co}~\cite{han2026vco} jointly denoises pixels and features with an \ttt{MMDiT}~\cite{esser2024scaling} architecture.
We employ \ttt{V-Co} as a joint image-feature denoising baseline and, to the best of our knowledge, are the first to study this paradigm for medical image generation.

\subsection{Medical-domain Generation and Transfer}

\noindent\textbf{Medical generative modelling.}
Diffusion and flow models are widely used for medical image generation. In 3D,
a \ttt{U-Net}~\cite{ronneberger2015unet} denoiser is typically trained in 3D \ttt{VQ-GAN}~\cite{esser2021taming} (discrete) or \ttt{VAE}~\cite{rombach2022high, kingma2014vae} (continuous) latents~\cite{khader2023denoising, muller2023multimodal},
directly in voxel space~\cite{xu2024medsyn,dorjsembe2024meddpm}, or in
wavelet space~\cite{friedrich2024wdm}.
At scale, \ttt{MAISI}~\cite{guo2025maisi} generates 3D \Th{CT} with a latent \Th{U-Net}, and \ttt{MAISIv2}~\cite{zhao2026maisi} reuses \ttt{MAISI}'s \ttt{VAE} and \Th{U-Net}, yet replaces the objective with rectified flow~\cite{liu2023flow}. \ttt{3D MedDiffusion}~\cite{wang20253d} pairs a custom VAE with a latent denoiser trained from scratch on medical data.
2D slices can also be stacked into 3D volumes~\cite{laslo2026dmg}.
In 2D, most works train the denoiser from
scratch~\cite{pinaya2022brain,moghadam2023morphology,pan2023transformer, huang2024chest,laslo2025mechanistic, liu2025treatmentaware}, yet a few fine-tune a pretrained text-to-image model, \eg, \ttt{RoentGen}~\cite{chambon2022roentgen}, \ttt{MediSyn}~\cite{cho2024medisyn}, and \ttt{Med-Art}~\cite{guo2025med}, fine-tune pretrained text-to-image
models.
In histopathology, \ttt{PathLDM}~\cite{yellapragada2024pathldm} initialises its
denoiser from \Th{IN} weights, improving generation, and \ttt{ZoomLDM}~\cite{yellapragada2025zoomldm} follows a similar recipe.
Most works treat transfer as an implementation detail.
We instead isolate the denoiser and study it systematically across size and paradigms.

\noindent\textbf{Transfer learning in medical imaging.}
Natural-image transfer is well studied in medical image analysis, where gains are often linked to faster convergence, scale, and feature reuse~\cite{ke2021chextransfer,raghu2019transfusion,mustafa2021supervised,matsoukas2022makes}. More recently, SSL natural-image encoders have also been transferred as initialisation or off-the-shelf for medical tasks, with competitive results compared to domain-specific models~\cite{wu2026braindino, karasikov2025midnight}. In contrast, we study transfer for generative modelling, focusing on whether natural-image denoisers transfer to medical generation and what transfers.

\section{Experimental Setup}
\label{sec:setup}

In this section, we describe our experimental setup, denoising schemes, datasets, evaluation protocol, initialisation schemes, and implementation details.

\subsection{Denoising models}
A denoiser $F(\cdot;\mathrm{W})$ with parameters $\mathrm{W}$ differs across paradigms by what it operates on. Pixel-space models denoise noisy images directly, $F(\mathbf{x}_t;\mathrm{W})$. Latent-space models first encode the image via a tokeniser, $\mathbf{z}=\mathcal{E}(\mathbf{x})$, and then denoise the noisy latent, $F(\mathbf{z}_t;\mathrm{W})$. The joint modelling approach employed in this study also operates in pixel space and denoises the noisy image together with an auxiliary semantic-feature stream, $\mathbf{s}=\Phi(\mathbf{x})$; we denote this as $F(\mathbf{x}_t,\mathbf{s};\mathrm{W})$; $t$ denotes the diffusion timestep/noise level.

\noindent \textbf{Latent denoising.}
We instantiate $F$ with \ttt{LDiT-XL}~\cite{yao2025reconstruction}, a $28$-layer ($675$M) latent diffusion transformer trained with a flow objective. We couple it with three tokenisers $\mathcal{E}$ that produce latents $\mathbf{z}$: i) \ttt{SDVAE}~\cite{rombach2022high}, the standard Stable Diffusion autoencoder trained on natural images; ii) \ttt{VAVAE}~\cite{yao2025reconstruction}, a natural-image VAE aligned with \ttt{DINOv2}~\cite{oquab2024dinov2}; and iii) \ttt{MedVAE}~\cite{varma2025medvae}, our medical-domain VAE baseline.

\noindent
\textbf{Pixel denoising.} We instantiate $F$ with \ttt{JiT}~\cite{li2026jit}, a diffusion transformer that operates directly on noisy pixels $\mathbf{x}_t$ and
utilises $\mathbf{x}$-prediction, and is trained via flow matching. We study three sizes, \ttt{B/16}, \ttt{L/16}, and \ttt{H/16} ($12$, $24$, $32$ layers; $131$M, $459$M, $952$M
parameters), which allows us to examine phenomena across models scales.

\noindent
\textbf{Joint denoising.} We instantiate $F$ with \ttt{V-Co-B/16}~\cite{han2026vco}, a joint denoising variant that operates on noisy pixels $\mathbf{x}_t$ together with an auxiliary \ttt{DINOv2} feature stream $\mathbf{s}=\Phi(\mathbf{x})$.

\subsection{Datasets}

\noindent
\textbf{\Th{RadImageNet}}~\cite{mei2022radimagenet}. is a 2D radiology dataset
covering CT and MR. Its CT and MR sub-collections comprise
$530{,}534$ slice PNGs spanning $5$
$(\text{modality},\text{body part})$ regions: CT abdomen, CT lung, MR
hip, MR knee, and MR shoulder. Each region is further subdivided into
per-pathology sub-folders, with 19 pathology classes
remaining after discarding the rare ones (fewer than $100$ samples) in our validation set.

\noindent
\textbf{\Th{BraTS}}~\cite{de2024brats24, menze2014multimodal}.
We use the 2024 BraTS post-treatment adult glioma cohort,
where each subject has four co-registered, skull-stripped MRI modalities
(\ttt{T1}, \ttt{T1c}, \ttt{T2}, \ttt{T2FLAIR}). From $1{,}621$ training and $188$
validation patients, we extract $226{,}940$ training and $26{,}320$ validation 4-channel samples. In the \emph{single-channel} setup, channel flattening yields $907{,}760$ training and $105{,}280$ validation images.
We detail more in the Supp.

\subsection{Spatial Structure Metrics}
\label{sec:ssms}
Spatial (structure) similarity metrics (SSMs) describe a visual entity by its internal geometric layout, \eg, how each patch resembles other patches of the same image, rather than by appearance such as colour or texture ~\cite{shechtman2007matching}.
Singh \etal \cite{singh2026what} show, that this property, of the \emph{target representation} (VFM),
predicts generation quality under representation alignment~\cite{yu2025representation}. We instead measure it on the \emph{internal representations} of the denoiser itself.

\noindent \textbf{LDS metric.}~\cite{huang1997lds}
Let $F_l(\mathbf{x}_t) \in \mathbb{R}^{N \times D}$ be the denoiser
representation at layer $l$ and noise level $t$, with $N$ patch tokens of
dimension $D$.
We measure self-similarity between tokens $i,j$
with the cosine kernel $K_{ij}$, and their spatial separation with the Manhattan distance $d_{ij}$. \Th{LDS}~\cite{huang1997lds} contrasts nearby ($R_{\text{n}}$) against distant ($R_{\text{f}}$) pairs,
\begin{equation*}
\mathrm{LDS} (F_l) = \mathrm{LDS}_l = \mathbb{E}_{d_{ij} < R_{\text{n}}} [K_{ij}]
             - \mathbb{E}_{d_{ij} \geq R_{\text{f}}} [K_{ij}],
\end{equation*}
where $i\neq j$, and $R_{\text{n}}, R_{\text{f}}$ are near and far thresholds.

\noindent \textbf{\LDSc metric.}
We modify LDS calculation by 1) mean-centering the tokens before computing the kernel and 2) excluding first and last blocks, whose representations remain close to the generation (pixel/latent) space, giving \LDSc (a centered version of LDS).

\subsection{Evaluation metrics}
\noindent
\textbf{Generation metrics.}
We report \Th{FID}~\cite{heusel2017gans} as our
primary generation-quality metric.
We additionally report density (\Th{Dens.}) and coverage
(\Th{Cov.})~\cite{naeem2020reliable}, more robust alternatives to precision and
recall~\cite{kynkaanniemi2019improved} that disentangle fidelity and diversity.
We detail both in the Supp.

\noindent
\textbf{Spectral profile of generated images.}
As a \emph{fine-grained diagnostic}, we compute the radial power spectrum of
generated images and report the per-frequency ratio
$P_{\text{gen}}(f)/P_{\text{real}}(f)$. A ratio of $1$ matches the real power at
that frequency band. This is a diagnostic, not a standalone metric. It is (more) informative for \emph{similar, within-family models}, \eg, the same \ttt{JiT} architecture under different learning rates, initialisations or sizes.

\subsection{Initialisation schemes}
\label{subsec:init}

\noindent
\textbf{Whole-model initialisation.}
We compare three schemes: 1) \textbf{\Th{IN}} loads the full \Th{IN}-init
denoiser weights; 2) \textbf{\Th{Random}} uses each model's default random
initialisation; and 3) \textbf{Statistically-matched} (\Th{ST})~\cite{raghu2019transfusion}
resamples each tensor from a Gaussian matching the per-tensor mean and variance
of the \Th{IN} weights, preserving the weight statistics but not the learned
structure. \Th{ST} thus isolates how much of the benefit comes from statistics
alone~\cite{raghu2019transfusion, matsoukas2022makes}.

\noindent
\textbf{Block reversion.}
Following~\cite{zhang2022all}, we take a \emph{fully fine-tuned} denoiser,
revert each transformer block to its original \Th{IN} weights and measure the
g\Th{FID} change. A large change marks a block which the model
relies on; a small one marks a block whose original \Th{IN} weights remain
useful.

\begin{table*}[!htb]
\centering
\caption{\textbf{RadImageNet (RadIN) 256$\times$256 gFID across the three denoising
paradigms.}
r\Th{FID} is the off-the-shelf tokenizer reconstruction \Th{FID};
\Th{Ep.}\ is the number of RadIN training epochs; \Th{Params} reports trainable
denoiser parameters only; $\Delta_{\text{init}}$ and $\Delta_{\text{init}}\%$ is the absolute and relative g\Th{FID}
improvement of \Th{IN} (\ch) vs \Th{Random} (\xmark) Denoiser, as indicated in \Th{Init} column.
\textbf{Bold} = best per category.}
\label{tab:radin-fid}
\footnotesize
\setlength{\tabcolsep}{5pt}
\begin{tabular}{cccccccccc}
\toprule
\Th{Denoiser} & \Th{Tokenizer} & \Th{Init.} & \Th{Ep.}
 & \Th{Params} & r\Th{FID}$\downarrow$ & g\Th{FID}$\downarrow$
 & \Th{Dens.}$\uparrow$ & \Th{Cov.}$\uparrow$
 & ${\Delta_{\text{init}}}$/${\Delta_{\text{init}}}(\%)$ \\
\midrule
\sectionrow{10}{\Th{Latent denoising}}
\ttt{LDiT-XL/2} & \cellcolor{medbg}\ttt{MedVAE} & \xmark & 80 & 675M & 6.35 & 21.33 & 0.64 & 0.72 & --- \\
\ttt{LDiT-XL/2} & \cellcolor{natbg}\ttt{SDVAE} & \xmark & 80 & 675M & 2.63 & 19.15 & 0.69 & 0.80 & --- \\
\ttt{LDiT-XL/1} & \cellcolor{natbg}\ttt{VAVAE} & \xmark & 80 & 675M & 1.88 & \underline{19.05} & \textbf{0.79} & \underline{0.85} & --- \\
\cellcolor{natbg}\ttt{LDiT-XL/1} & \cellcolor{natbg}\ttt{VAVAE} & \ch & 80 & 675M & 1.88 & \textbf{16.43} & \underline{0.77} & \textbf{0.86} & 2.6 / 13.8 \\
\midrule
\sectionrow{10}{\Th{Pixel denoising}}
\ttt{JiT-B/16} & --- & \xmark & 80 & 131M & --- & 55.50 & 0.27 & 0.31 & --- \\
\cellcolor{natbg}\ttt{JiT-B/16} & --- & \ch & 80 & 131M & --- & 36.48 & 0.39 & 0.51 & 19.0 / 34.3 \\
\ttt{JiT-L/16} & --- & \xmark & 80 & 459M & --- & 41.27 & 0.44 & 0.50 & --- \\
\cellcolor{natbg}\ttt{JiT-L/16} & --- & \ch & 80 & 459M & --- & \textbf{32.34} & \textbf{0.48} & \textbf{0.60} & 8.9 / 21.6 \\
\ttt{JiT-H/16} & --- & \xmark & 40 & 952M & --- & 47.89 & 0.30 & 0.34 & --- \\
\cellcolor{natbg}\ttt{JiT-H/16} & --- & \ch & 40 & 952M & --- & \underline{32.83} & \underline{0.46} & \underline{0.60} & 15.1 / 31.4 \\
\midrule
\sectionrow{10}{\Th{Joint denoising}}
\cellcolor{natbg}\ttt{V-Co-B/16} & \cellcolor{natbg}\ttt{DINOv2} & \ch & 80 & 260M & --- & 28.86 & 0.44 & 0.56 & --- \\
\bottomrule
\end{tabular}
\\[4pt]
\footnotesize
\colorbox{medbg}{\phantom{XX}} Medical-image model \quad
\colorbox{natbg}{\phantom{XX}} Natural-image model
\end{table*}

\subsection{Implementation details}

\noindent
\textbf{Models and training.}
We use the pretrained \ttt{VAVAE}, \ttt{SDVAE}, and \ttt{MedVAE} tokenisers off
the shelf. As \ttt{VAVAE} applies higher spatial compression, we pair it with
\ttt{LDiT-XL/1} (latent patch size $1$), and the lower-compression \ttt{SDVAE}
and \ttt{MedVAE} with \ttt{LDiT-XL/2} (patch size $2$). The denoisers are our
object of study: we follow each repository's recipe and hyperparameters for
\ttt{LDiT}\footnote{\url{https://github.com/hustvl/LightningDiT}}~\cite{yao2025reconstruction}, \ttt{JiT}\footnote{\url{https://github.com/LTH14/JiT}}~\cite{li2026jit}, and
\ttt{V-Co}\footnote{\url{https://github.com/HL-hanlin/V-Co}}~\cite{han2026vco}, with pixel models \ttt{JiT-\{B,L,H\}/16} and
\ttt{V-Co-B/16} using a $16{\times}16$ patch size.
All models are trained at $256{\times}256$ resolution, with \Th{RadIN} models conditioned on $5$ body-part classes and \Th{BraTS} models on $2$ classes (tumour/no tumour). We train for $80$ epochs on \Th{RadIN}, $40$ epochs on single-channel \Th{BraTS}, and $55$ epochs on multi-channel \Th{BraTS}, unless otherwise stated. Additional details are provided in Supp.

\noindent
\textbf{Sampling.}
We follow each model's recommended sampler and step budget:
\ttt{LDiT} uses Euler ODE sampler with $250$ steps, while
\ttt{JiT} and \ttt{V-Co} use Heun with $50$ steps. We use
$\text{cfg}{=}1.0$ throughout to measure transfer effects without confounding from guidance tuning.

\noindent
\textbf{Evaluation.}
We compute all metrics with the \ttt{dgm-eval}\footnote{\url{https://github.com/layer6ai-labs/dgm-eval}}
suite~\cite{stein2023exposing}, reporting g\Th{FID}, \Th{Dens} , and \Th{Cov}.
We evaluate against each dataset’s held-out test split: $15$k images for \Th{RadIN} and $50$k for \Th{BraTS}. This split is not used for training or per-model hyperparameter tuning.

\section{Natural-image transfer improves medical image generation}
\label{sec:exps}

\subsection{The case of RadImageNet}
\label{subsec:radin}

\autoref{tab:radin-fid} reports generation metrics (g\Th{FID}, \Th{Dens}, \Th{Cov}) on
\Th{RadIN} across the three denoising paradigms.

\noindent \textbf{Transfer helps, and most in pixel space.}
\Th{IN} transfer consistently improves generation, but its magnitude differs
across paradigms. In latent space, \ttt{LDiT-XL/1} improves by $13.8\%$, whereas in pixel space \ttt{JiT} improves by $34.3\%$
(\ttt{B}), $21.6\%$ (\ttt{L}), and $31.4\%$ (\ttt{H}), with density and
coverage following suit. Latent thus attains the best absolute performance
($16.43$), while pixel gains most from transfer ($21$-$34\%$). The tokeniser likely
explains this gap: natural-image \Th{VAE}s compress images spatially $8\text{-}16\times$ and
reconstruct radiology well (1.88 r\Th{FID} \ttt{VAVAE}) despite no medical training, giving
the denoiser 1) much of the transferable prior already, and 2) an easier,
lower-dimensional problem. Together, these explain the latent denoiser's strong
performance (${\sim}12$ g\Th{FID} over pixel/joint) and why \Th{IN}-init adds a smaller yet significant, gain ($2.62$ g\Th{FID}). Pixel
diffusion has no tokeniser: the denoiser benefits from the prior directly, so
transfer helps more. The joint \ttt{V-Co} (28.86
g\Th{FID}) falls in the same performance regime. We therefore focus on pixel-space
\ttt{JiT}, where transferability lies in the denoiser.

\noindent \textbf{Larger capacity does not uniformly improve RadIN gFID.}
Within the \ttt{JiT} family, scaling from \ttt{B/16} to \ttt{L/16} improves g\Th{FID} by over $4$ points, and density and coverage by $\sim$$0.1$. Scaling to \ttt{H/16} ($952$M),
however, does not improve g\Th{FID}. Thus, in our setup and data regime ($\sim$$0.5$M images), increasing capacity beyond \ttt{L/16} does not translate into a clear gain in generation metrics.

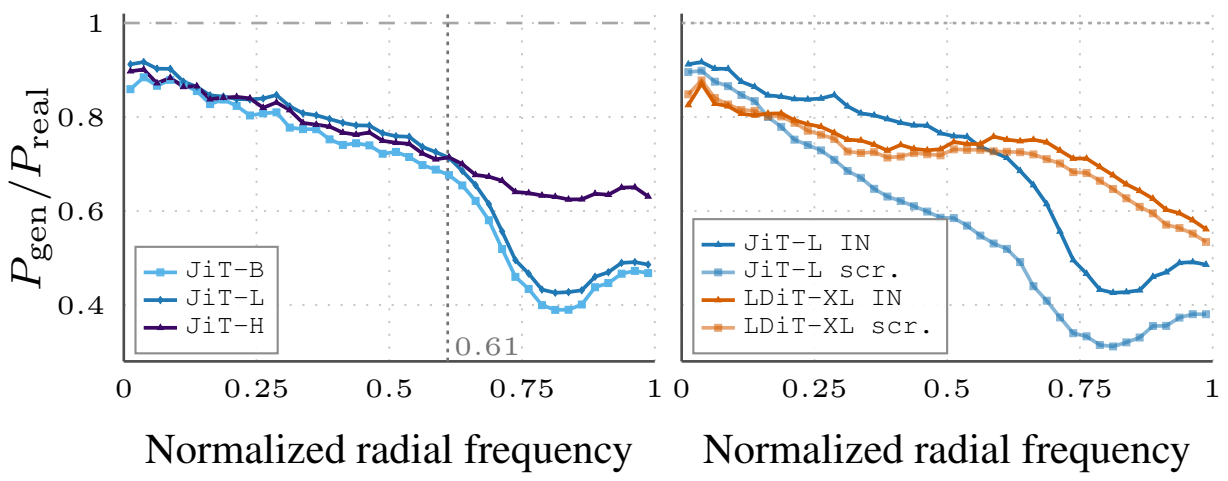
\begin{figure}[!htb]
  \centering
  \definecolor{jitB}{HTML}{56B4E9}
  \definecolor{jitL}{HTML}{1F77B4}
  \definecolor{jitH}{HTML}{3B0764}
  \definecolor{lditC}{HTML}{D55E00}
  \pgfplotsset{
    specaxis/.style={
      scale only axis,
      width=0.86\linewidth,
      height=2.3cm,
      font=\footnotesize,
      tick label style={font=\tiny}, label style={font=\scriptsize},
      legend style={font=\tiny, fill=white, fill opacity=0.9, text opacity=1,
                    draw=black!45, inner sep=1.2pt, row sep=-2pt,
                    nodes={inner sep=1.2pt}},
      legend cell align=left,
      legend image code/.code={%
        \draw[mark repeat=2, mark phase=2]
          plot coordinates {(0cm,0cm) (0.11cm,0cm) (0.22cm,0cm)};%
      },
      xmin=0, xmax=1, ymin=0.28, ymax=1.02,
      xtick={0,0.25,0.5,0.75,1.0}, ytick={0.4,0.6,0.8,1.0},
      grid=major, grid style={dotted, gray!45, line width=0.35pt},
      axis line style={line width=0.5pt, black!70},
      tick style={draw=none},
      axis x line*=bottom, axis y line*=left,
      xlabel near ticks, ylabel near ticks,
      ylabel shift={-6pt},
      xlabel={Normalized radial frequency},
      ylabel={$P_\mathrm{gen}/P_\mathrm{real}$},
      every axis plot/.append style={line width=0.7pt, mark size=0.65pt,
                                     mark options={line width=0.2pt}},
    },
  }
  \begin{subfigure}[t]{0.49\columnwidth}
    \centering
    \begin{tikzpicture}
      \begin{axis}[specaxis, legend style={at={(0.02,0.02)}, anchor=south west}]
        \addplot[gray!70, dashed, line width=0.5pt, mark=none, forget plot,
                 domain=0:1, samples=2] {1};
        \addplot[jitB, mark=square*] coordinates {
        (0.0125,0.8592) (0.0375,0.8845) (0.0625,0.8665) (0.0875,0.8791) (0.1125,0.8694)
        (0.1375,0.8552) (0.1625,0.8275) (0.1875,0.8371) (0.2125,0.8237) (0.2375,0.8033)
        (0.2625,0.8077) (0.2875,0.8099) (0.3125,0.7775) (0.3375,0.7742) (0.3625,0.7740)
        (0.3875,0.7524) (0.4125,0.7406) (0.4375,0.7441) (0.4625,0.7396) (0.4875,0.7219)
        (0.5125,0.7257) (0.5375,0.7149) (0.5625,0.6977) (0.5875,0.6881) (0.6125,0.6763)
        (0.6375,0.6544) (0.6625,0.6216) (0.6875,0.5803) (0.7125,0.5191) (0.7375,0.4599)
        (0.7625,0.4341) (0.7875,0.3996) (0.8125,0.3896) (0.8375,0.3899) (0.8625,0.4010)
        (0.8875,0.4378) (0.9125,0.4466) (0.9375,0.4666) (0.9625,0.4725) (0.9875,0.4678)};
          \addlegendentry{\ttt{JiT-B}}
        \addplot[jitL, mark=diamond*] coordinates {
        (0.0125,0.9122) (0.0375,0.9170) (0.0625,0.9028) (0.0875,0.9025) (0.1125,0.8751)
        (0.1375,0.8643) (0.1625,0.8461) (0.1875,0.8435) (0.2125,0.8382) (0.2375,0.8375)
        (0.2625,0.8394) (0.2875,0.8468) (0.3125,0.8227) (0.3375,0.8079) (0.3625,0.8037)
        (0.3875,0.7957) (0.4125,0.7877) (0.4375,0.7823) (0.4625,0.7818) (0.4875,0.7656)
        (0.5125,0.7592) (0.5375,0.7579) (0.5625,0.7364) (0.5875,0.7254) (0.6125,0.7133)
        (0.6375,0.6854) (0.6625,0.6551) (0.6875,0.6148) (0.7125,0.5565) (0.7375,0.4955)
        (0.7625,0.4667) (0.7875,0.4327) (0.8125,0.4260) (0.8375,0.4274) (0.8625,0.4313)
        (0.8875,0.4600) (0.9125,0.4697) (0.9375,0.4896) (0.9625,0.4913) (0.9875,0.4859)};
          \addlegendentry{\ttt{JiT-L}}
        \addplot[jitH, mark=triangle*] coordinates {
        (0.0125,0.8974) (0.0375,0.9011) (0.0625,0.8720) (0.0875,0.8834) (0.1125,0.8640)
        (0.1375,0.8657) (0.1625,0.8378) (0.1875,0.8411) (0.2125,0.8429) (0.2375,0.8397)
        (0.2625,0.8196) (0.2875,0.8310) (0.3125,0.8147) (0.3375,0.7873) (0.3625,0.7838)
        (0.3875,0.7797) (0.4125,0.7667) (0.4375,0.7625) (0.4625,0.7671) (0.4875,0.7494)
        (0.5125,0.7452) (0.5375,0.7429) (0.5625,0.7224) (0.5875,0.7100) (0.6125,0.7142)
        (0.6375,0.6998) (0.6625,0.6768) (0.6875,0.6730) (0.7125,0.6644) (0.7375,0.6406)
        (0.7625,0.6381) (0.7875,0.6330) (0.8125,0.6303) (0.8375,0.6246) (0.8625,0.6249)
        (0.8875,0.6367) (0.9125,0.6346) (0.9375,0.6498) (0.9625,0.6505) (0.9875,0.6309)};
          \addlegendentry{\ttt{JiT-H}}
        \addplot[black!55, densely dotted, line width=0.5pt, mark=none, forget plot]
          coordinates {(0.61,0.28) (0.61,1.02)};
        \node[anchor=south west, font=\tiny, text=black!55, inner sep=1pt]
          at (axis cs:0.61,0.28) {$0.61$};
      \end{axis}
    \end{tikzpicture}
  \end{subfigure}
  \hfill
  \begin{subfigure}[t]{0.49\columnwidth}
    \centering
    \begin{tikzpicture}
      \begin{axis}[specaxis, ylabel={}, yticklabels={},
                   legend style={at={(0.02,0.02)}, anchor=south west}]
        \addplot[gray!70, densely dotted, line width=0.5pt, mark=none, forget plot,
                 domain=0:1, samples=2] {1};
        \addplot[jitL, mark=triangle*] coordinates {
        (0.0125,0.9122) (0.0375,0.9170) (0.0625,0.9028) (0.0875,0.9025) (0.1125,0.8751)
        (0.1375,0.8643) (0.1625,0.8461) (0.1875,0.8435) (0.2125,0.8382) (0.2375,0.8375)
        (0.2625,0.8394) (0.2875,0.8468) (0.3125,0.8227) (0.3375,0.8079) (0.3625,0.8037)
        (0.3875,0.7957) (0.4125,0.7877) (0.4375,0.7823) (0.4625,0.7818) (0.4875,0.7656)
        (0.5125,0.7592) (0.5375,0.7579) (0.5625,0.7364) (0.5875,0.7254) (0.6125,0.7133)
        (0.6375,0.6854) (0.6625,0.6551) (0.6875,0.6148) (0.7125,0.5565) (0.7375,0.4955)
        (0.7625,0.4667) (0.7875,0.4327) (0.8125,0.4260) (0.8375,0.4274) (0.8625,0.4313)
        (0.8875,0.4600) (0.9125,0.4697) (0.9375,0.4896) (0.9625,0.4913) (0.9875,0.4859)};
          \addlegendentry{\ttt{JiT-L IN}}
        \addplot[jitL, opacity=0.55, mark=square*] coordinates {
        (0.0125,0.8956) (0.0375,0.8984) (0.0625,0.8752) (0.0875,0.8656) (0.1125,0.8469)
        (0.1375,0.8338) (0.1625,0.8001) (0.1875,0.7791) (0.2125,0.7518) (0.2375,0.7404)
        (0.2625,0.7290) (0.2875,0.7090) (0.3125,0.6852) (0.3375,0.6707) (0.3625,0.6472)
        (0.3875,0.6307) (0.4125,0.6212) (0.4375,0.6101) (0.4625,0.5988) (0.4875,0.5857)
        (0.5125,0.5849) (0.5375,0.5689) (0.5625,0.5477) (0.5875,0.5308) (0.6125,0.5194)
        (0.6375,0.4912) (0.6625,0.4399) (0.6875,0.4091) (0.7125,0.3735) (0.7375,0.3399)
        (0.7625,0.3333) (0.7875,0.3148) (0.8125,0.3118) (0.8375,0.3204) (0.8625,0.3307)
        (0.8875,0.3553) (0.9125,0.3556) (0.9375,0.3732) (0.9625,0.3804) (0.9875,0.3803)};
          \addlegendentry{\ttt{JiT-L scr.}}
        \addplot[lditC, mark=triangle*] coordinates {
        (0.0125,0.8255) (0.0375,0.8708) (0.0625,0.8283) (0.0875,0.8231) (0.1125,0.8069)
        (0.1375,0.8034) (0.1625,0.8072) (0.1875,0.8078) (0.2125,0.7932) (0.2375,0.7847)
        (0.2625,0.7792) (0.2875,0.7664) (0.3125,0.7517) (0.3375,0.7501) (0.3625,0.7411)
        (0.3875,0.7287) (0.4125,0.7407) (0.4375,0.7323) (0.4625,0.7292) (0.4875,0.7315)
        (0.5125,0.7469) (0.5375,0.7426) (0.5625,0.7425) (0.5875,0.7584) (0.6125,0.7518)
        (0.6375,0.7486) (0.6625,0.7518) (0.6875,0.7462) (0.7125,0.7287) (0.7375,0.7116)
        (0.7625,0.7117) (0.7875,0.6938) (0.8125,0.6761) (0.8375,0.6569) (0.8625,0.6428)
        (0.8875,0.6263) (0.9125,0.6027) (0.9375,0.5953) (0.9625,0.5805) (0.9875,0.5613)};
          \addlegendentry{\ttt{LDiT-XL IN}}
        \addplot[lditC, opacity=0.55, mark=square*] coordinates {
        (0.0125,0.8483) (0.0375,0.8779) (0.0625,0.8402) (0.0875,0.8267) (0.1125,0.8150)
        (0.1375,0.8132) (0.1625,0.8017) (0.1875,0.8012) (0.2125,0.7877) (0.2375,0.7708)
        (0.2625,0.7624) (0.2875,0.7534) (0.3125,0.7266) (0.3375,0.7234) (0.3625,0.7253)
        (0.3875,0.7137) (0.4125,0.7158) (0.4375,0.7233) (0.4625,0.7208) (0.4875,0.7184)
        (0.5125,0.7312) (0.5375,0.7302) (0.5625,0.7303) (0.5875,0.7270) (0.6125,0.7260)
        (0.6375,0.7251) (0.6625,0.7204) (0.6875,0.7104) (0.7125,0.7011) (0.7375,0.6827)
        (0.7625,0.6810) (0.7875,0.6644) (0.8125,0.6468) (0.8375,0.6270) (0.8625,0.6089)
        (0.8875,0.5951) (0.9125,0.5709) (0.9375,0.5637) (0.9625,0.5519) (0.9875,0.5342)};
          \addlegendentry{\ttt{LDiT-XL scr.}}
      \end{axis}
    \end{tikzpicture}
  \end{subfigure}
    \caption{
  \textbf{Radial power-spectrum ratio}:
    $P_{\mathrm{gen}}(f)/P_{\mathrm{real}}(f)$ on \Th{RadIN}. A ratio of 1 means the generated samples match real power at that frequency.}
  \label{fig:spectrum_ratio}
\end{figure}

\noindent \textbf{Spectral diagnostics show differences in generated image structure.}
In \autoref{fig:spectrum_ratio} we use the spectral profile of generated images as a within-family diagnostic of image structure, across three comparisons. First, across \ttt{JiT} sizes: all models reproduce low-to-mid frequencies more closely than high ones, but while \ttt{B/16} and \ttt{L/16} undershoot high-frequency power, \ttt{H/16} better preserves it. \ttt{L/16} is in turn slightly better aligned in the low-to-mid bands, which may explain its better g\Th{FID}; so although \ttt{H/16} does not improve g\Th{FID} on \Th{RadIN}, it does change the frequency profile. Second, \Th{Random} vs \Th{IN}-initialised \ttt{JiT-L/16} shows that natural-image initialisation alters the generated spectrum. Finally, in latent space the difference between \Th{Random} and \Th{IN} \ttt{LDiT-XL/1} is smaller, consistent with the tokeniser already providing a structured representation.

\subsection{The case of BraTS}
\label{subsec:brats}

We next study transfer on BraTS, in single and multi-channel scenarios.

\begin{table}[!htb]
\centering
\caption{\textbf{BraTS-GLI 256$\times$256 g\Th{FID} across single- and
multi-channel diffusion.}
\textbf{(a)} Single-channel: the denoiser randomly generates \emph{one} of the
four MRI modalities;
\textbf{(b)} Multi-channel: the denoiser generates
\emph{all four} MRI modalities;
\Th{Ep.}\ the number of training epochs; \textbf{Bold} = best per panel.}
\label{tab:brats-fid}
\footnotesize
\begin{minipage}[t]{0.49\textwidth}
\centering
\textbf{(a) Single-channel}
\setlength{\tabcolsep}{4pt}
\begin{tabular}{lcccccc}
\toprule
\Th{Denoiser} & \Th{Init.} & \Th{Ep.}
 & g\Th{FID}$\downarrow$ & \Th{Dens.}$\uparrow$ & \Th{Cov.}$\uparrow$ & ${\Delta_{\text{init}}}$ \\
\midrule
\ttt{LDiT-XL/1} & \xmark & 40 & 9.22 & 0.67 & 0.76 & --- \\
\cellcolor{natbg}
\ttt{LDiT-XL/1} & \ch    & 40 & \textbf{7.66} & \textbf{0.68} & \textbf{0.78} & 1.56 \\
\midrule
\ttt{JiT-L/16}   & \xmark & 40 & 37.93 & 0.19 & 0.26 & ---\\
\cellcolor{natbg}
\ttt{JiT-L/16}   & \ch    & 40 & \underline{29.34} & \textbf{0.25} & \textbf{0.36} & 8.59 \\
\cellcolor{natbg}
\ttt{JiT-H/16}   & \ch    & 10 & \textbf{29.13} & \underline{0.23} & \underline{0.35} & --- \\
\bottomrule
\end{tabular}
\label{tab:brats_single_channel}
\end{minipage}%
\\[4pt]
\begin{minipage}[t]{0.49\textwidth}
\centering
\textbf{(b) Multi-channel}
\setlength{\tabcolsep}{4pt}
\begin{tabular}{lccccc}
\toprule
\Th{Denoiser} & \Th{Init.} & \Th{Ep.}
 & g\Th{FID}$\downarrow$ & \Th{Dens.}$\uparrow$ & \Th{Cov.}$\uparrow$ \\
\midrule
\ttt{JiT-L/16} & \xmark & 55 & 43.97 & 0.09 & 0.14 \\
\midrule
\cellcolor{natbg}
\ttt{JiT-L/16} & \ch    & 55 & 35.17 & 0.15 & 0.23 \\
\midrule
\cellcolor{natbg}
\ttt{JiT-H/16} & \ch    & 55 & \textbf{32.75} & \textbf{0.18} & \textbf{0.26} \\
\bottomrule
\end{tabular}
\end{minipage}
\end{table}

\noindent
\textbf{Single-channel MRI reproduces the transfer pattern.}
\Th{IN} initialisation improves both latent and pixel-space models:
\ttt{LDiT-XL/1} by $1.56$ g\Th{FID} and \ttt{JiT-L/16} by $8.59$, again a larger
absolute gain in pixel space. \ttt{JiT-H/16} reaches comparable g\Th{FID} to
\ttt{JiT-L/16} in fewer epochs ($10$ vs.\ $40$). Since \ttt{H/16} has
roughly twice the FLOPs of \ttt{L/16},\footnote{At fixed sequence length,
transformer training FLOPs are approximately proportional to the parameter
count, so doubling the parameters roughly doubles the FLOPs.} this corresponds
to about half the compute.

\noindent
\textbf{Multi-channel generation benefits from capacity.}
Generating all four modalities jointly is more challenging than the single-channel task:
\ttt{JiT-L/16} g\Th{FID} rises from $29.34$ to $35.17$. \Th{IN} initialisation
nonetheless remains beneficial, improving pretrained \ttt{JiT-L/16} by $8.80$
g\Th{FID} over \Th{Random}. Unlike \Th{RadIN}, additional capacity
here improves all metrics, with \ttt{JiT-H/16} outperforming \ttt{JiT-L/16},
suggesting capacity may matter more when jointly representing multiple
co-registered MRI contrasts.

\subsection{Do FID gains carry clinical signal?}
To assess whether FID improvement translates to clinically relevant information we perform the following experiment. We first sample 100k samples from our training set.
We then train two \ttt{JiT-B/16} (scratch; IN) models on the whole \Th{RadIN}, conditioned on all (67) classes, and generate 100k samples over the 19 (validation) classes. This way we get three 100k pools with identical pathology distributions.

\begin{table}[!htb]
\centering
\footnotesize
\caption{\textbf{Clinical Proxies on synthetic data}. We measure \Th{FD-Rad} and \Th{macroF1} on two synthetic 100k pools.}
\label{tab:tstr}
\begin{tabular}{cccccc}
\toprule
\Th{Denoiser} & \Th{Init.} & gFID$\downarrow$ & \Th{FD-rad}$\downarrow$ & \Th{macroF1}$\uparrow$ \\
\midrule
\cellcolor{natbg}
\ttt{JiT-B/16}  & \ch      & \bf{40.81} & \textbf{37.99} & \textbf{0.47} \\
\ttt{JiT-B/16}  & \xmark  & 72.67 & 87.46 & 0.43 \\
\bottomrule
\end{tabular}
\end{table}
\noindent
\textbf{\Th{FD-Rad}} In \autoref{tab:tstr}, we use 93 radiomic features~\cite{pyrad}, a radiology feature space capturing low-level image properties and measure the Fréchet Distance from the real (100k) radiomic distribution. We find that lower gFID translates to lower \Th{FD-Rad}, \ie, more clinically meaningful images.

\noindent
\textbf{Pathology Classification} We train
\ttt{ResNet-50} models on real and synthetic data as a proxy to clinical (19-class pathology) classification.
The classifier trained on the real data pool achieves 0.72.
For the synthetic pools (\autoref{tab:tstr}) we find that lower gFID translates to better \Th{macroF1}, pointing towards more meaningful clinical generation.

\section{What transfers: spatial structure, not semantics}
\label{sec:analysis}
Since the pixel denoiser carries the entire prior, we focus on pixel-space \ttt{JiT} and ask whether the benefit reflects high-level (semantics) or low-level (spatial) structure.

\begin{figure}[!htb]
  \centering
    \includegraphics[width=0.99\columnwidth]{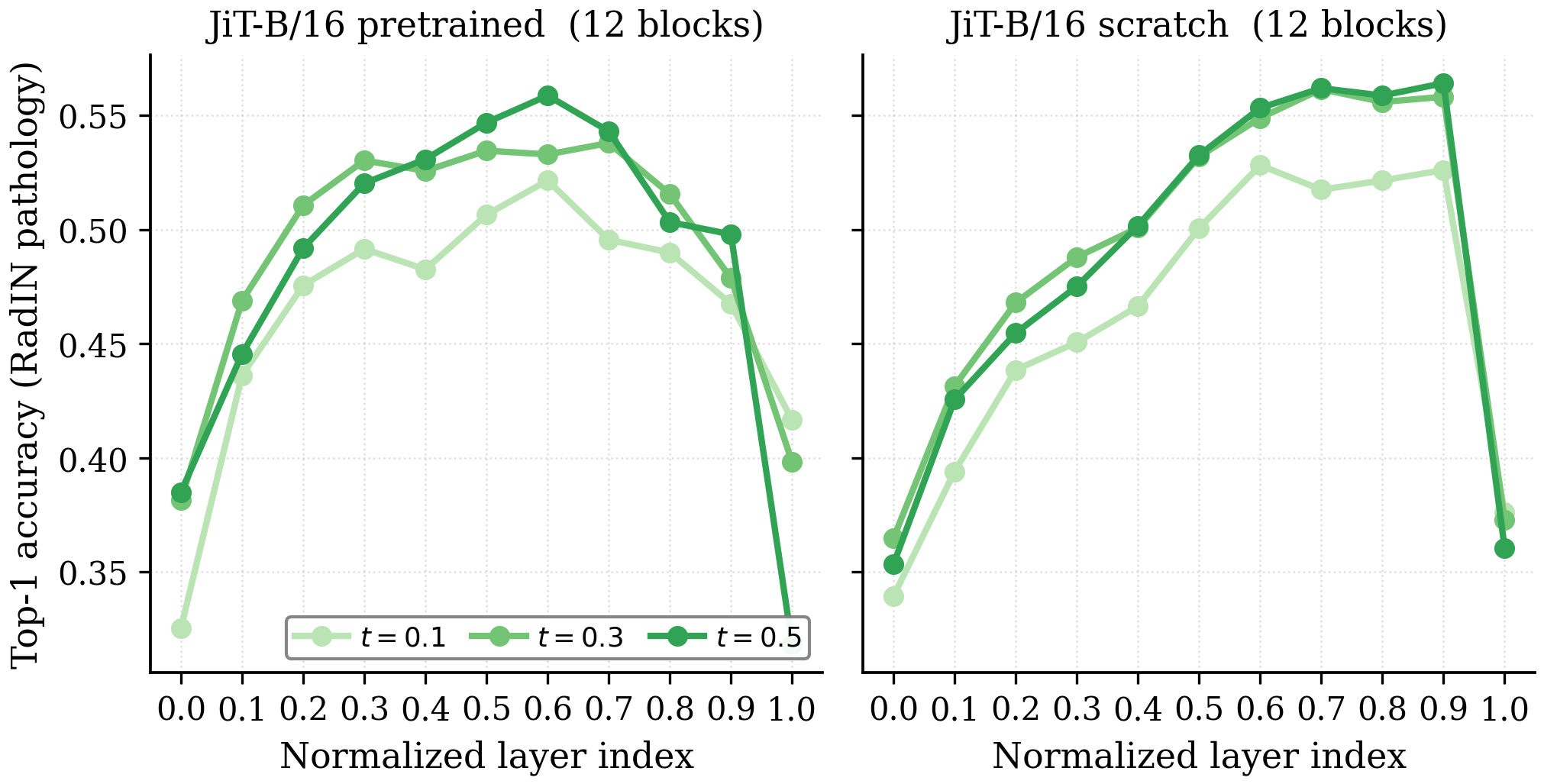}
\caption{\textbf{Semantic probing.} Top-1 accuracy of linear probes on frozen
\ttt{JiT-B} representations classifying $19$ \Th{RadIN} pathology classes, by
layer depth and noise level $t$. Averages over three runs.}
\label{fig:linear_probing}
\end{figure}

\subsection{Semantics do not predict generation quality}
Following~\cite{yu2025representation}, a \emph{hypothesis} is that pretraining helps via \emph{semantics}, either by transferring semantic
features or by providing an initialisation from which it learns stronger semantics. We probe frozen denoiser representations to classify the $19$ pathology classes of \Th{RadIN}. Since the generative models are conditioned on $5$ broad anatomy classes, probes capture semantic information beyond generation labels.
\par
\emph{Higher probe accuracy does not translate to better generation.}
\ttt{JiT-B/16} from scratch probes similar to its IN counterpart
($0.56$) yet generates far worse samples ($55.50$ vs $36.48$
g\Th{FID}, \autoref{fig:linear_probing}). The same holds across model sizes (\ttt{L/H} see Supp.), with all models performing comparably.
\ttt{JiT-B/16} probes slightly higher than \ttt{JiT-H/16}
($0.56$ vs $0.54$) but has worse g\Th{FID} ($36.48$ vs $32.83$). In latent
space the scratch model also probes higher than the \Th{IN}-init
($0.557$ vs $0.510$) while generating worse (Supp.).
Semantic separability cannot explain the transfer benefit.

\subsection{Spatial structure tracks generation quality}
\label{subsec:structure}
\begin{figure}[!htb]
\centering
\begin{tikzpicture}
\begin{axis}[
    width=0.7\linewidth,
    height=0.48\linewidth,
    xlabel={Spatial Structure (\LDSc)  $\longrightarrow$},
    ylabel={$\longleftarrow$  gFID},
    xlabel style={font=\footnotesize},
    ylabel style={font=\footnotesize},
    xmin=0.60, xmax=0.665,
    ymin=29, ymax=50,
    xtick={0.60,0.61,0.62,0.63,0.64,0.65,0.66},
    ytick={30,35,40,45,50},
    tick label style={font=\scriptsize},
    grid=major,
    grid style={line width=0.2pt, draw=black!12},
    axis line style={draw=black!55},
    axis lines*=left,
    legend style={
        at={(1.03,0.5)}, anchor=west,
        font=\scriptsize,
        cells={anchor=west},
        legend columns=2,
        row sep=0.1em,
        draw=black!35,
        fill=white,
        fill opacity=0.95,
        text opacity=1,
    },
    legend cell align={left},
    every axis plot post/.append style={mark options={draw=white, line width=0.7pt}},
]
\addplot[dashed, thick, red!70, domain=0.60:0.665, samples=2, forget plot]
    {225.62 - 296.56*x};
\addplot[only marks, mark=*, mark size=2.8pt, color=blue!85!black]
    coordinates {(0.6327, 36.48)};                   \addlegendentry{B$_{\rm IN\;s0}$}
\addplot[only marks, mark=*, mark size=2.8pt, color=blue!70!black]
    coordinates {(0.6340, 35.55)};                   \addlegendentry{B$_{\rm IN\;s1}$}
\addplot[only marks, mark=*, mark size=2.8pt, color=blue!45!black]
    coordinates {(0.6232, 37.90)};                   \addlegendentry{B$_{\rm IN\;50\%}$}
\addplot[only marks, mark=*, mark size=2.8pt, color=blue!25!black]
    coordinates {(0.6104, 42.50)};                   \addlegendentry{B$_{\rm IN\;25\%}$}
\addplot[only marks, mark=*, mark size=3.8pt, color=blue!55!cyan]
    coordinates {(0.6570, 32.34)};                   \addlegendentry{L$_{\rm IN}$}
\addplot[only marks, mark=square*, mark size=3.8pt, color=red!75!black]
    coordinates {(0.6394, 41.27)};                   \addlegendentry{L$_{\rm scratch}$}
\addplot[only marks, mark=*, mark size=4.5pt, color=cyan!65!white]
    coordinates {(0.6463, 32.83)};                   \addlegendentry{H$_{\rm IN}$}
\addplot[only marks, mark=square*, mark size=4.5pt, color=red!35!white!85!black]
    coordinates {(0.6197, 47.89)};                   \addlegendentry{H$_{\rm scratch}$}
\addplot[only marks, mark=triangle*, mark size=3.9pt, color=green!45!black]
    coordinates {(0.6561, 31.46)};                   \addlegendentry{L$_{\rm mix5}$}
\addplot[only marks, mark=triangle*, mark size=3.9pt, color=green!60!black]
    coordinates {(0.6546, 30.94)};                   \addlegendentry{L$_{\rm mix15}$}
\addplot[only marks, mark=triangle*, mark size=3.9pt, color=green!70!black]
    coordinates {(0.6529, 31.01)};                   \addlegendentry{L$_{\rm mix25}$}
\addplot[only marks, mark=triangle*, mark size=3.9pt, color=green!45!white]
    coordinates {(0.6464, 31.80)};                   \addlegendentry{L$_{\rm mix50}$}
\node[anchor=north east, font=\small,
      fill=white, fill opacity=0.95, text opacity=1,
      draw=black!35, rounded corners=1.5pt, inner sep=3pt]
      at (rel axis cs:0.98,0.96)
      {$|r| = 0.846$};
\end{axis}
\end{tikzpicture}
\caption{
    \textbf{\LDSc correlation with gFID across 12 \ttt{JiT}
    variants.} \LDSc here is the max value across \ttt{JiT} layers, at $t{=}0.5$.
}
\label{fig:lds_gfid_scatter}
\end{figure}
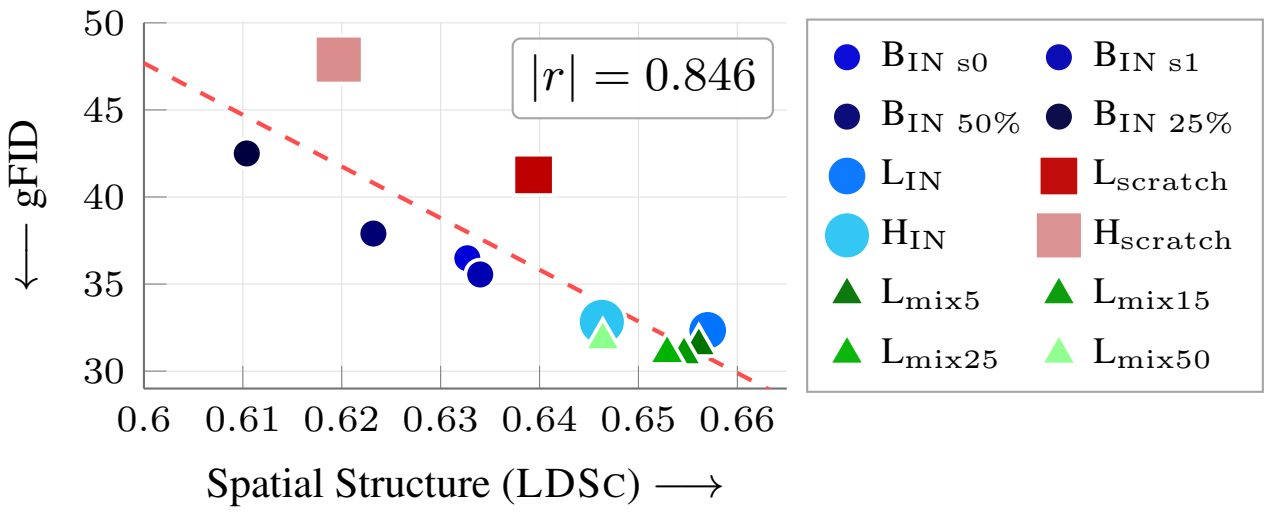

An alternative \emph{hypothesis} is that the benefit comes from \emph{spatial} (low-level) structure~\cite{matsoukas2022makes, singh2026what}. To test this, we collect all \ttt{JiT} models trained in this paper, \ie, pretrained, scratch, mixed (\autoref{sec:mixed}), and limited-data (\autoref{sec:limited_data}). Applying LDS (\autoref{sec:ssms}) to frozen denoiser representations gives inconclusive results. We modify LDS for denoiser representations, giving \LDSc, and track gFID at $|r|{=}0.846$ over 12 models (\autoref{fig:lds_gfid_scatter}). The fitted law is
$\Delta\text{gFID}{=}{-}296.6 \times \Delta\LDSc$. We ask how much of the pretrained-scratch gFID gap this rule predicts from the
\LDSc gap alone. We multiply the observed \LDSc gap by the fitted slope to obtain a predicted gFID gap, and divide by the actual gFID gap to get the fraction that the \LDSc signal captures. For \ttt{B/L/H} it captures $92/58/52\%$, or $67\%$ on average. Two checks confirm this: 1) Monte-Carlo
perturbation of every measured gFID by $\sigma\!=\!1$ (more than the seed noise), refitting over $5$k runs gives $69 \pm 6\%$; 2) Removing both IN and scratch models of a
size before refitting and predicting that size's gap gives $95/62/42\%$, again 2/3 on average. \LDSc
predicts about two-thirds of the pretraining benefit. \emph{Better spatial organisation is one channel through which pretraining manifests.}

\begin{figure*}[!bht]
  \centering
  \includegraphics[width=0.85\linewidth]{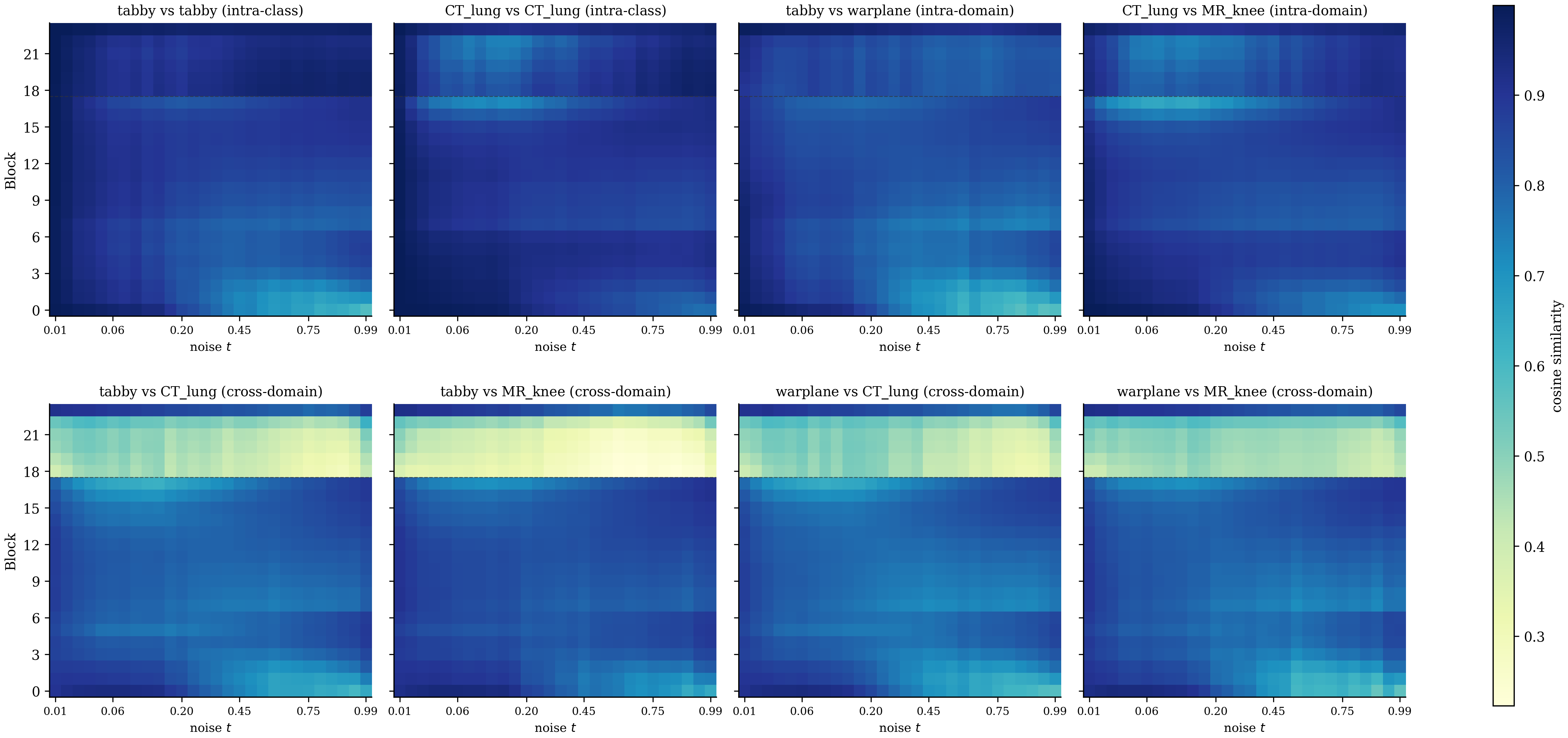}
  \caption{%
    \textbf{Cosine similarity of internal representations in mixed-JiT-L/16.}
    Blocks (y-axis) and noise levels $t$ (x-axis). Intra-class and intra-domain pairs (top row);
    Cross-domain pairs (bottom row).}
  \label{fig:class_aware_cond}
\end{figure*}

\subsection{Mixed-domain training}
\label{sec:mixed}

\noindent \textbf{Early-to-mid representations are shared across domains}
We further test the structural hypothesis and train \ttt{JiT-L/16} jointly on
\Th{RadIN} and $10\%$ \Th{IN} (\ttt{mixed-JiT-L/16}), so a single model
represents both domains internally, and measure the per-block cosine similarity
of its representations (concatenated over the token sequence) under natural and
medical inputs. \autoref{fig:class_aware_cond} shows that early-to-mid blocks
($0$--$16$) stay highly similar across domains ($\cos\!\geq\!0.7$), so the model computes similar representations for both domains there. Deeper blocks
($17$--$22$) behave differently: cross-domain similarity drops
($\cos\!\approx\!0.35$--$0.50$), suggesting domain-specific behaviour.

\begin{table}[!htb]
\centering
\footnotesize
\caption{\textbf{Natural-image mixing.} g\Th{FID} of \ttt{JiT-L/16} (IN) trained for 80 epochs on mixed medical and natural data.}
\label{tab:mixtures}
\begin{tabular}{lccccc}
\toprule
IN mix (\%) & 0 & 5 & 15 & 25 & 50 \\
RadIN:IN & 1:0 & 19:1 & 5.7:1 & 3:1 & 1:1 \\
\midrule
gFID$\downarrow$ & 32.34 & 31.46 & \textbf{30.94} & 31.01 & 31.80 \\
\bottomrule
\end{tabular}
\end{table}

\noindent \textbf{Natural-image mixing helps medical fidelity}
Since both domains yield similar representations, we sweep the ratio of natural images added to the mix. g\Th{FID} improves across all ratios (\autoref{tab:mixtures}), with the largest gain ($1.4$ gFID) at $15\%$. Even at 1:1 we observe a small improvement, though its magnitude sits near gFID noise.

\section{The benefit is tied to natural-image initialisation}
\label{sec:where}

We ask whether the benefit of \Th{IN} initialisation can be reproduced without it, or \emph{decomposed} into simpler parts.
To reproduce it, we test in-domain medical pretraining, statistically matched reinitialisation, and medical data scaling; none matches \Th{IN} transfer in our setup.
To decompose it, we compare parameter drift with block reversion, and find that drift does not predict functional importance.

\subsection{Natural-image transfer outperforms medical pretraining}
A natural alternative to \Th{IN}-init is to pretrain on medical data. To our knowledge, no
2D radiology denoiser weights exist at a compatible architecture, so we train our own.

\noindent \textbf{Single-channel target}
A \ttt{JiT-B/16} pretrained from scratch on \Th{RadIN} for 160 epochs and fine-tuned on single-channel \Th{BraTS} for 20 epochs is $5$ g\Th{FID} worse than direct \Th{IN} fine-tuning (\autoref{tab:e3_pretraining_regime}). A \Th{RadIN} stage on top of \Th{IN} helps slightly, but within seed noise ($\sigma_{\text{IN}}{=}0.41$, Supp.).
\begin{table}[h]
\centering
\caption{
\textbf{Single-channel BraTS transfer}.
}
\label{tab:e3_pretraining_regime}
\footnotesize
\setlength{\tabcolsep}{4pt}
\begin{tabular}{lccccc}
\toprule
\Th{Denoiser (init)} & \Th{Pretrain} & \Th{Ep.}
 & g\Th{FID}$\downarrow$ & \Th{Dens.}$\uparrow$ & \Th{Cov.}$\uparrow$ \\
\midrule
\ttt{JiT-B/16} \Th{(Rand)} & \cellcolor{medbglight}\Th{RadIN} & 160 & 36.73 & 0.209 & 0.277 \\
\midrule
\cellcolor{natbg}
\ttt{JiT-B/16} (\Th{IN}) & \cellcolor{medbglight}\Th{RadIN} & 80 & \textbf{31.13} & \textbf{0.221} & 0.320 \\
\cellcolor{natbg}
\ttt{JiT-B/16} (\Th{IN}) & --- & --- & 31.73 & 0.218 & \textbf{0.322} \\
\bottomrule
\end{tabular}
\end{table}

\noindent \textbf{Multi-channel target.}
We test whether an intermediate single-channel medical stage helps before
transferring to multi-channel \Th{BraTS} (\autoref{tab:brats-routes}). Direct
\Th{IN} transfer is the best ($40.17$). Inserting a medical stage only
hurts, giving $50.90$ after single-channel \Th{BraTS} and $75.40$ after \Th{RadIN}.

\begin{table}[h]
\centering
\caption{\textbf{Multi-channel BraTS transfer}}
\label{tab:brats-routes}
\footnotesize
\centering
\setlength{\tabcolsep}{4pt}
\begin{tabular}{lccc}
\toprule
\Th{Denoiser} & \Th{Pretrain} & \Th{Ep.}
 & g\Th{FID}$\downarrow$ \\
\midrule
\cellcolor{natbg}
\ttt{JiT-L/16} & \cellcolor{medbglight}\Th{RadIN} & 30 & 75.40\\
\cellcolor{natbg}
\ttt{JiT-L/16} & \cellcolor{medbg}\Th{BraTS} & 30 & 50.90\\
\midrule
\cellcolor{natbg}
\ttt{JiT-L/16} & --- & 30 & \textbf{40.17}\\
\bottomrule
\end{tabular}
\end{table}

\subsection{Per-tensor weight statistics are not sufficient}
If \Th{IN} only set a suitable per-tensor scale, statistically matched (\Th{ST}) initialisation would recover its benefit. Instead, \Th{ST} is \emph{worse} than \Th{Random} in both paradigms. \autoref{tab:init_schemes} shows g\Th{FID} rises from $41.27$ to $62.64$ for \ttt{JiT-L} and from $19.05$ to $51.45$ for \ttt{LDiT-XL}. The benefit comes from the specific weight values, not their distribution.

\begin{table}[h]
  \centering
  \caption{\textbf{Initialisation schemes.} g\Th{FID} on \Th{RadIN} after $80$ epochs from
\Th{Random}, \Th{ST}, and \Th{IN} initialisation across both
paradigms.}
  \label{tab:init_schemes}
  \footnotesize
  \begin{tabular}{l c c c}
    \toprule
    \Th{Denoiser} & \Th{Random} & \Th{ST} & \Th{IN} \\
    \midrule
    \ttt{LDiT-XL/1} & 19.05& 51.45 & \textbf{16.43}\\
    \midrule
    \ttt{JiT-L/16} & 41.27& 62.64 & \textbf{32.34}\\
    \bottomrule
  \end{tabular}
\end{table}

\subsection{Parameter drift and functional importance}
\label{sec:drift-scale}
We ask whether a layer's change during training reflects its importance. We measure change with \emph{parameter drift} and importance with \emph{block reversion}.

\begin{figure}[!htb]
  \centering
  \begin{subfigure}[t]{0.49\columnwidth}
    \centering
    \includegraphics[width=\linewidth]{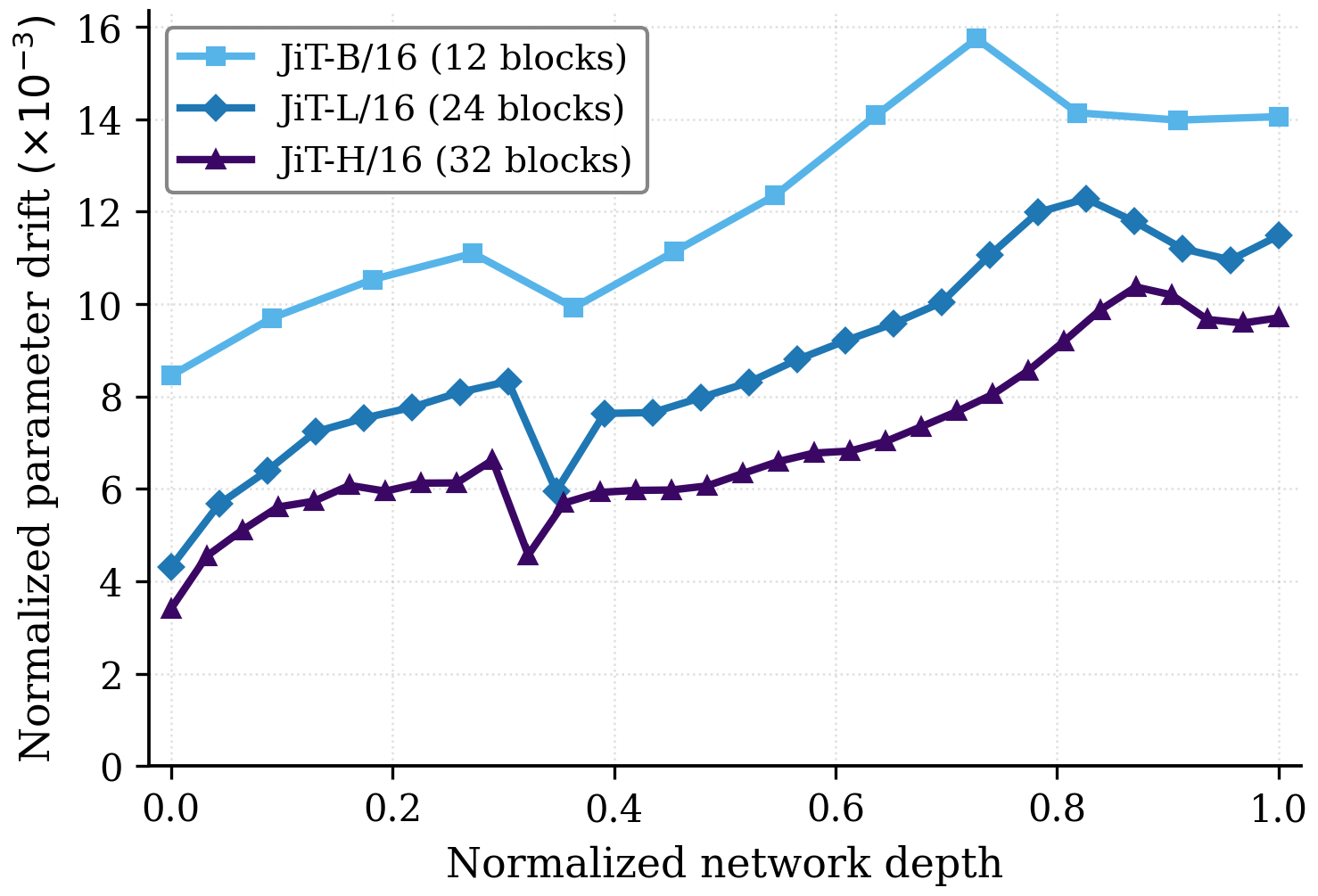}
  \end{subfigure}
  \hfill
  \begin{subfigure}[t]{0.49\columnwidth}
    \centering
    \includegraphics[width=\linewidth]{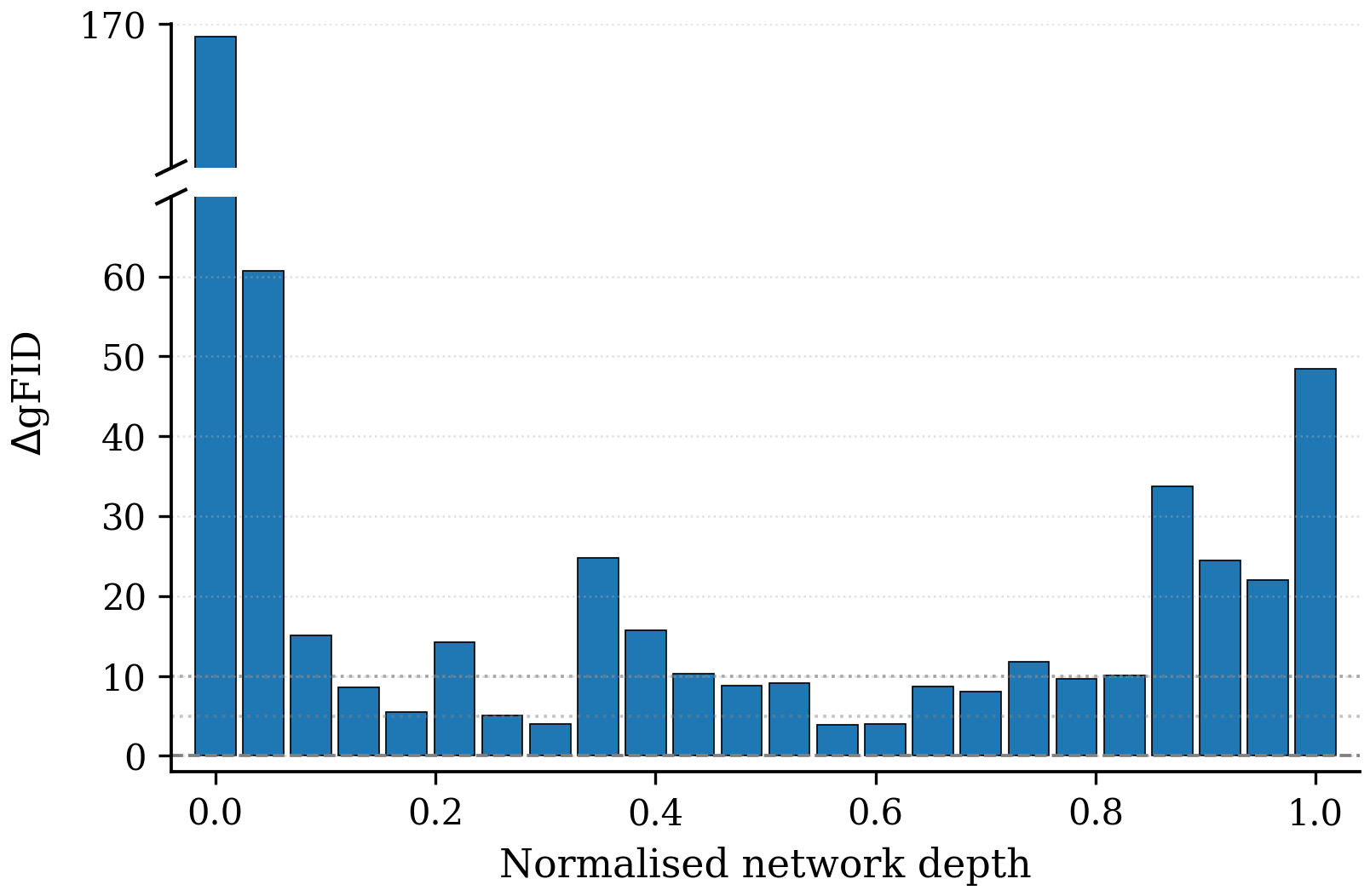}
  \end{subfigure}
  \caption{\Th{(Left)} \textbf{Per transformer block normalized parameter drift}:
  $\|\Delta W\|_F / \sqrt{N_{\text{block}}}$ from \Th{IN} init. All models trained for 20 epochs. \Th{(Right)} \textbf{Per-block reversion importance}: $\Delta$gFID for
each block of \ttt{JiT-L/16}.
}
  \label{fig:drift_reinit}
\end{figure}

\noindent \textbf{Larger models drift less.}
\autoref{fig:drift_reinit} \Th{(Left)} shows that drift increases with depth. Early
blocks remain closer to their \Th{IN}-init, while deeper blocks move further. Averaged over blocks, drift decreases with model
size: $12.10$ (\ttt{B}), $8.80$ (\ttt{L}), and $6.98$ (\ttt{H}) (${\times}10^{-3}$), so
larger models adapt less.

\noindent \textbf{Functional importance is not captured by drift alone.}
Block reversion (\autoref{fig:drift_reinit} \Th{Right}) gives a different picture from parameter drift.
Reverting the \emph{first} transformer block to its \Th{IN} weights is by
far the most damaging, raising g\Th{FID} by $\sim$$170$, even though this block
drifts \emph{less} than the deeper ones. The second and four final blocks are also
important. Several intermediate blocks can be reverted with smaller
changes in g\Th{FID}.
To account for any activation mismatch we also perform activation renormalisation after block reversion in the Supp.
Results follow similar trend, thus drift cannot capture functional importance.

\subsection{The gap is not closed by data scaling}
\label{sec:limited_data}
A final hypothesis is that more medical data lets the model learn from scratch what transfer provides.

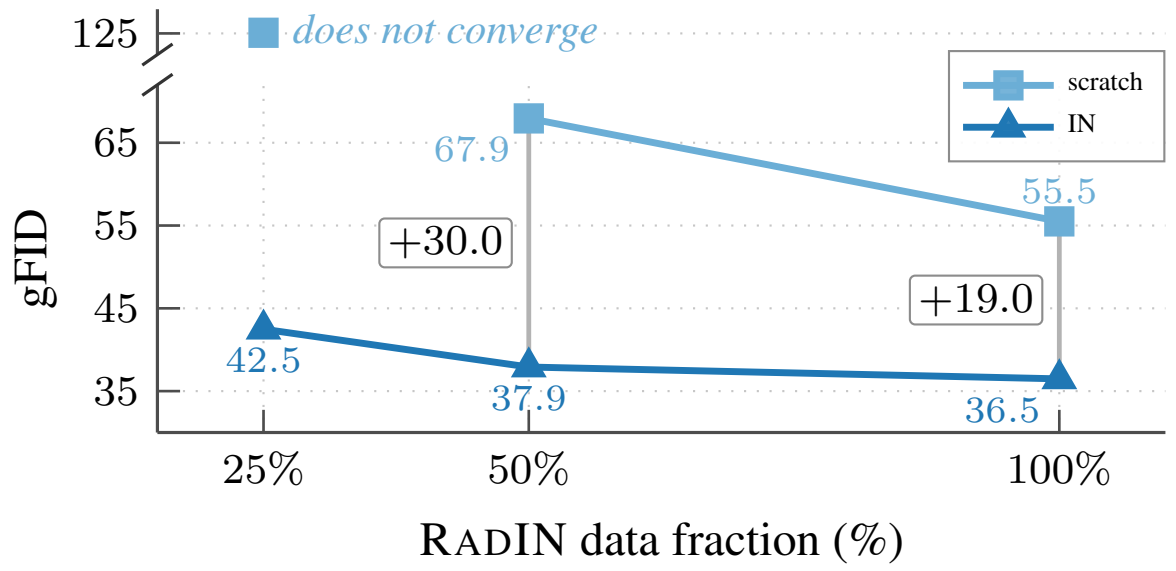
\begin{figure}[t]
  \centering
  \definecolor{preC}{HTML}{1F77B4}
  \definecolor{scrC}{HTML}{6BAED6}
  \newcommand{\dsann}[1]{{\fontsize{7}{8}\selectfont #1}}
 \pgfplotsset{
    dsaxis/.style={
      scale only axis,
      width=0.8\linewidth,
      font=\small,
      tick label style={font=\footnotesize}, label style={font=\small},
      legend style={font=\footnotesize, fill=white, fill opacity=0.92,
                    text opacity=1, draw=black!45, inner sep=2pt,
                    row sep=-1pt, nodes={inner sep=2pt}},
      legend cell align=left,
      grid style={dotted, gray!45, line width=0.4pt},
      axis line style={line width=0.6pt, black!70},
      tick style={line width=0.5pt, black!70},
      axis x line*=bottom, axis y line*=left,
      xlabel near ticks, ylabel near ticks,
      every axis plot/.append style={line width=1.3pt},
    },
  }
  \begin{tikzpicture}
    \begin{axis}[dsaxis,
      name=dtop, height=0.3cm,
      xmin=15, xmax=110, ymin=118, ymax=133, ytick={125},
      axis x line=none, xlabel={}, ylabel={},
      ymajorgrids=true, xmajorgrids=false,
    ]
      \addplot[scrC, only marks, mark=square*, mark size=2.2pt]
        coordinates {(25,125.21)};
      \node[anchor=west, text=scrC, font=\footnotesize, xshift=2pt]
        at (axis cs:25,125.21) {\textit{does not converge}};
    \end{axis}

    \begin{axis}[dsaxis,
      name=dbot, at={(dtop.south west)}, anchor=north west, yshift=-2mm,
      height=2.3cm,
      xmin=15, xmax=110, ymin=30, ymax=72,
      xtick={25,50,100}, xticklabels={$25\%$,$50\%$,$100\%$},
      ytick={35,45,55,65},
      xlabel={\Th{RadIN} data fraction (\%)},
      ylabel={gFID},
      legend style={at={(1.,1.1)}, anchor=north east},
      ymajorgrids=true, xmajorgrids=true,
    ]
      \addplot[gray!60, line width=0.9pt, mark=none, forget plot]
        coordinates {(50,37.91) (50,67.94)};
      \addplot[gray!60, line width=0.9pt, mark=none, forget plot]
        coordinates {(100,36.48) (100,55.50)};
      \node[anchor=east, draw=black!45, fill=white, rounded corners=1pt,
            inner sep=1.6pt, font=\footnotesize, xshift=-3pt]
        at (axis cs:50,52.9) {$+30.0$};
    \node[anchor=east, draw=black!45, fill=white, rounded corners=1pt,
          inner sep=1.6pt, font=\footnotesize, xshift=-3pt]
      at (axis cs:100,46.0) {$+19.0$};

      \addplot[scrC, mark=square*, mark size=2.2pt]
        coordinates {(50,67.94) (100,55.50)};
        \addlegendentry{\tiny{scratch}}
      \addplot[preC, mark=triangle*, mark size=2.6pt]
        coordinates {(25,42.48) (50,37.91) (100,36.48)};
        \addlegendentry{\tiny{IN}}

      \node[anchor=north east,      text=scrC] at (axis cs:50,67.94)  {\dsann{$67.9$}};
      \node[anchor=south , text=scrC] at (axis cs:100,55.50) {\dsann{$55.5$}};
      \node[anchor=north,      text=preC] at (axis cs:25,42.48)  {\dsann{$42.5$}};
      \node[anchor=north,      text=preC] at (axis cs:50,37.91)  {\dsann{$37.9$}};
      \node[anchor=north east, text=preC] at (axis cs:100,36.48) {\dsann{$36.5$}};
    \end{axis}

    \draw[line width=0.6pt, black!80]
      ($(dtop.south west)+(-1mm,-0.7mm)$) -- ($(dtop.south west)+(1mm,0.7mm)$);
    \draw[line width=0.6pt, black!80]
      ($(dbot.north west)+(-1mm,-0.7mm)$) -- ($(dbot.north west)+(1mm,0.7mm)$);
  \end{tikzpicture}
    \caption{\textbf{Data scaling.} g\Th{FID} of \ttt{JiT-B/16} trained $80$ epochs
from scratch versus \Th{IN} on $25\%$, $50\%$, and $100\%$ of
\Th{RadIN}.}
    \label{fig:b16_data_scaling}
\end{figure}

\noindent \textbf{Limited data analysis}
We train \ttt{JiT-B/16} from both scratch and \Th{IN} init on $25\%$, $50\%$, and $100\%$ of
\Th{RadIN} (\autoref{fig:b16_data_scaling}). The pretrained model leads at every scale, the gap narrows (from $30.0$ to $19.0$) but stays large. IN on $25\%$ outperforms scratch on 100\% ($42.5$ vs $55.5$), while scratch fails to converge at $25\%$.

\noindent \textbf{Data requirements extrapolation}
Fitting $\text{FID}{\propto}N^{-\alpha}$ to each model, gives $\alpha_{\small{\text{Rand}}}=0.29$ and $\alpha_{\text{IN}}=0.11$.
If scratch improves at $\alpha_{\text{Rand}}$ until it reaches the
fidelity of \Th{IN} at $25\%$, and at $\alpha_{\text{IN}}$ thereafter, it needs about $10\times$
more data to close the gap (an illustrative estimate).

\section{Conclusion \& Discussion}
\label{sec:discussion}
We present a systematic study of natural-image denoiser transfer for medical image generation.
\Th{IN} transfer consistently improves generation performance across latent and pixel space, model
scales, and two radiology datasets.
Gains are also mirrored in radiomic distance and pathology classification.
Models trained from scratch reach similar semantic probe accuracy yet generate worse samples, whereas spatial organisation, measured by \LDSc{}, correlates
with gFID at $|r|{=}0.85$, positioning spatial structure as a channel through which pretraining
manifests. Although natural and medical images differ in high-level content, a mixed-domain denoiser computes similar representations for both through most of the network, and mixing natural images into medical training improves fidelity. The benefit is tied to natural-image initialisation, and no alternative we test reproduces it, including in-domain medical pretraining. From a generative modelling lens, natural images are very useful for medical image generation.

\noindent
\textbf{Limitations and future work.}
\LDSc predicts about two-thirds of the pretraining benefit, so spatial organisation is one channel through which it manifests, but might not be the full account. What exactly is inherited (edges, textures or contours), and how architecture, data, training budget and optimisation contribute, remains open.
Moreover, whether the benefits of natural images can be learned from medical data alone remains an open question.
Our extrapolation scales data quantity, not diversity.
Furthermore, our clinical evidence relies on proxies, and expert clinical evaluation is needed before any claim of clinical utility.
Finally, we study 2D generation, the native format for X-ray and histopathology, in which even volumetric scans are read slice by slice. Extending to 3D generation is a natural next step.

\section{Acknowledgements}
This work was supported by the Swiss National Science Foundation (TMSGI3\_225913, CRSK-3\_229103) and the Chad Tough Defeat DIPG Foundation.
We acknowledge EuroHPC Joint Undertaking for awarding us access to Discoverer+ (project EHPC-DEV-2026D06-289), hosted by PetaSC at Sofia Tech Park, Bulgaria, and co-funded by EuroHPC JU and the Bulgarian government.
E. Georgiou would like to thank T. Kouzelis for connecting him with I. Kakogeorgiou.

{
    \small
    \bibliographystyle{ieeenat_fullname}
    \bibliography{main}
}

\section{Additional Results}

\begin{table}[!htb]
\centering
\small
\caption{\textbf{rFID of off-the-shelf Tokenizers on BraTS and RadImageNet (RadIN) ($256{\times}256$).} The medical-domain \ttt{MedVAE} is dominated by every natural-image VAE on both datasets, despite the latter having seen no medical data.}
\label{tab:rfid}
\begin{tabular}{lcc}
\toprule
 & \multicolumn{2}{c}{r\Th{FID} $\downarrow$} \\
\cmidrule(lr){2-3}
\Th{Tokenizer} & \Th{BraTS} & \Th{RadIN} \\
\midrule
\rowcolor{natbg!60}
\multicolumn{3}{c}{\Th{Natural-image}} \\
\ttt{SDVAE}                       & 2.27 & 2.63 \\
\ttt{SD3.5VAE}                    & 0.48 & 0.92 \\
\ttt{VAVAE}                       & 1.34 & 1.88 \\
\ttt{RAEv2}\,(\ttt{DINOv3-L-k7})  & 1.60 & 2.42 \\
\addlinespace[2pt]
\rowcolor{medbg!60} \multicolumn{3}{c}{\Th{Medical-domain}} \\
\ttt{MedVAE}                      & 6.43 & 6.35 \\
\bottomrule
\end{tabular}
\end{table}

\noindent
Theoretically, an optimal compressor approaches the Kolmogorov complexity of its training data: it learns patterns rather than memorising, and reuses them to compress OOD data. We find natural VAEs compress radiology images very well with no medical training.
We treat this as another positive transfer result, and
\emph{hypothesise} MedVAE's performance stems from limited data and optimisation, e.g, training epochs, regularisation. We do not claim that medical data and pretraining do not help in general. Our illustrative extrapolation suggests that closing the gap would need far more medical data, whereas natural images are abundant and transfer to radiology.

\begin{table}[!htb]
\centering
\small
\caption{\textbf{Seed variance.} g\Th{FID} on \Th{RadIN} after $80$ epochs.
$\delta$ is the range (max$-$min) and $\sigma$ the standard deviation across seeds. \Th{IN} seeds share a fixed initialisation and vary only in data order, noise, etc.,
which likely explains why $\sigma_{\text{Rand}} > \sigma_{\text{IN}}$.}
\label{tab:seeds}
\begin{tabular}{lcccc}
\toprule
& \multicolumn{2}{c}{\ttt{JiT-B/16}} & \multicolumn{2}{c}{\ttt{LDiT-XL/1}} \\
\cmidrule(lr){2-3}\cmidrule(lr){4-5}
\Th{Seed} & \Th{Rand.} & \Th{IN} & \Th{Rand.} & \Th{IN} \\
\midrule
$0$      & 55.50 & 36.48 & 19.05 & 16.43 \\
$1$      & 56.30 & 35.71 & 18.62 & 16.28 \\
$2$      & 54.27 & 36.33 & ---   & ---   \\
\midrule
\Th{Mean} & 55.36 & 36.17 & 18.84 & 16.36 \\
$\delta$  & 2.03  & 0.77  & 0.43  & 0.15  \\
$\sigma$  & 1.02  & 0.41  & -  & -  \\
\bottomrule
\end{tabular}
\end{table}

\par

\begin{figure}[!htb]
\begin{subfigure}{0.85\columnwidth}
    \centering
    \includegraphics[width=\linewidth]{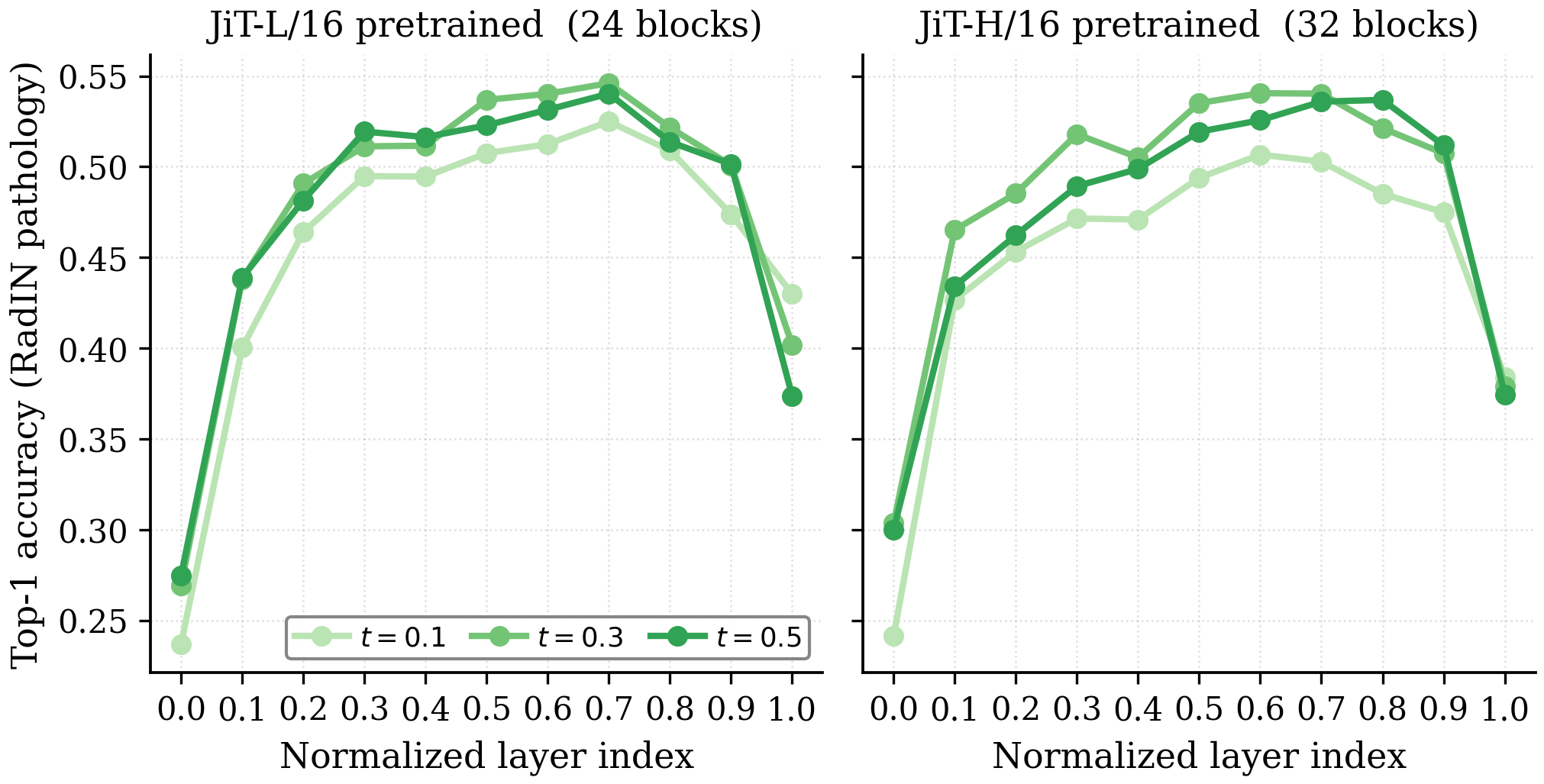}
    \caption{\ttt{JiT-L/16}; \ttt{JiT-H/16} (pixel-space)}
    \label{fig:jit_l_h_16}
\end{subfigure}
\begin{subfigure}{0.85\columnwidth}
    \centering
    \includegraphics[width=\linewidth]{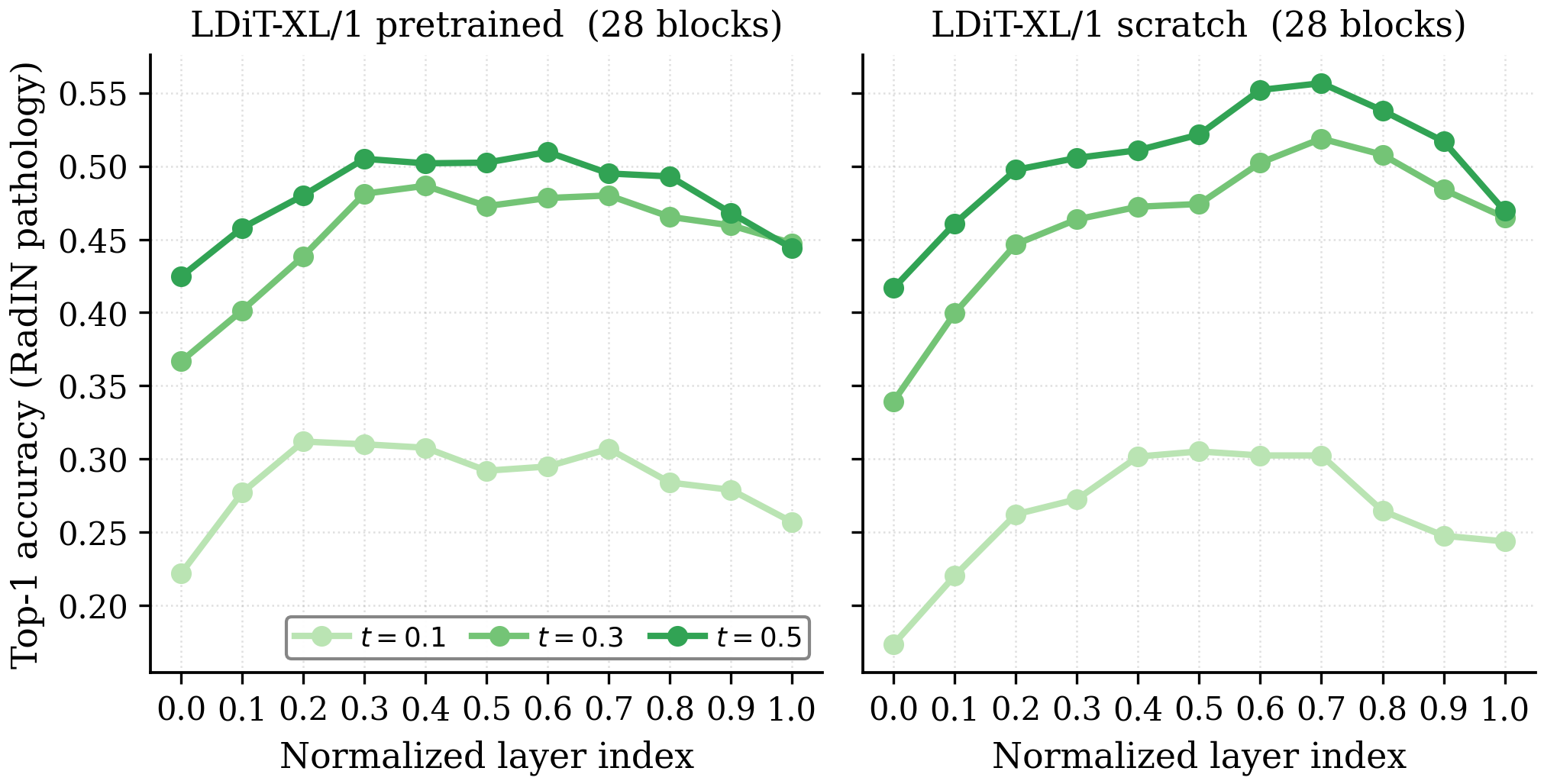}
    \caption{\ttt{LDiT-XL/1}; left pretrained; right scratch.}
    \label{fig:ldit_xl_1}
\end{subfigure}
\caption{\textbf{Semantic/pathology probing results}.}
\end{figure}

\par

\begin{figure}[!htb]
\centering
\includegraphics[width=0.9\linewidth]{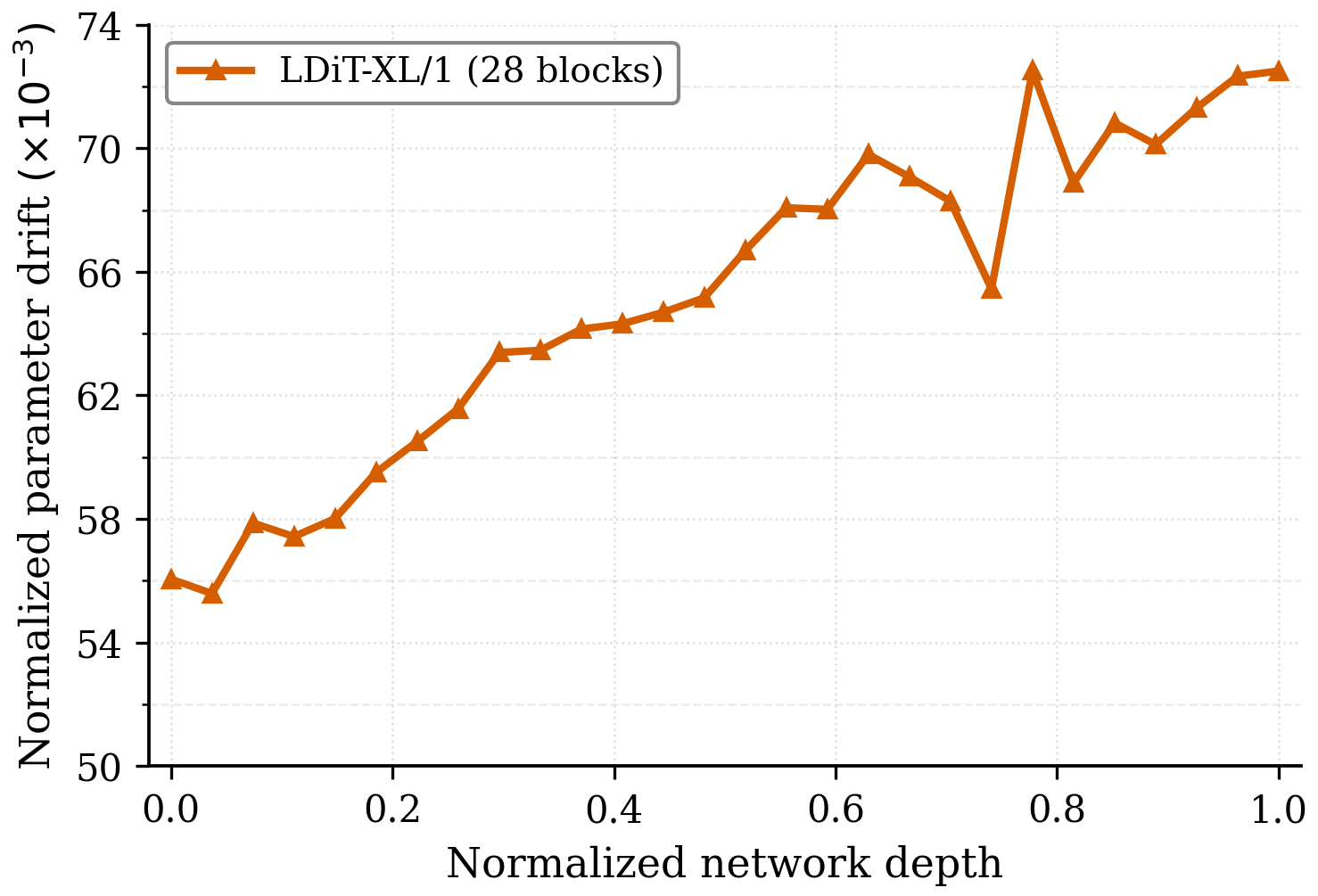}
\caption{\textbf{Parameter drift of \Th{IN}-init \ttt{LDiT-XL/1}}}
\end{figure}

\par

\begin{figure}[!htb]
\centering
\includegraphics[width=0.9\linewidth]{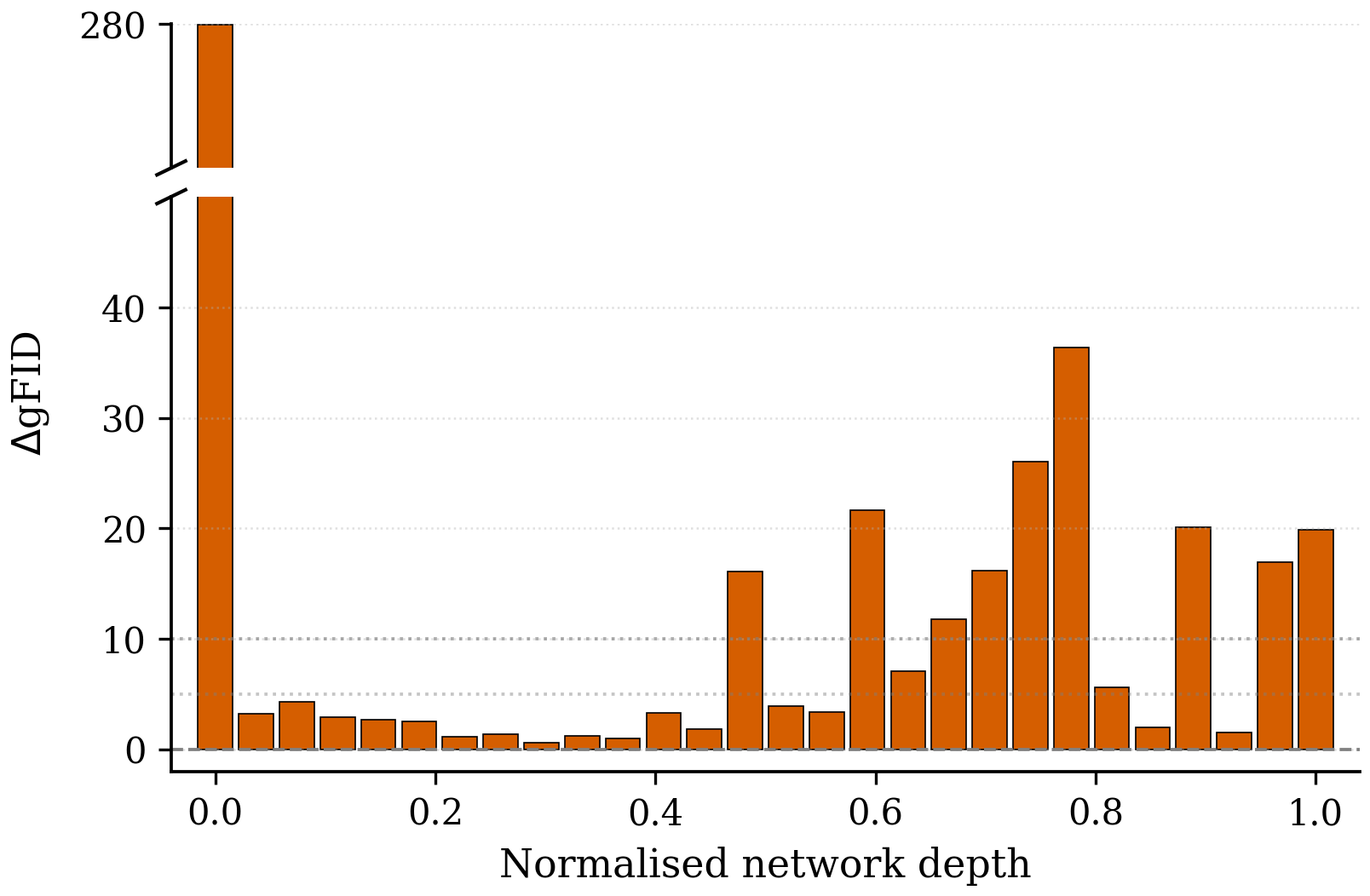}
\caption{\textbf{Per-block reversion of \ttt{LDiT-XL/1}}}
\end{figure}

\par

\begin{table}[!htb]
\centering
\footnotesize
\setlength{\tabcolsep}{5pt}
\caption{\textbf{Block reversion with activation renormalisation.}
$\Delta$g\Th{FID} ($\uparrow$ worse) on \Th{RadIN} when a single block $L_k$ of the
fine-tuned \ttt{JiT-L/16} is reverted to the IN weights. \Th{+ Renorm} additionally rescales the block $k$ output to match the original
per-channel statistics $(\mu, \sigma)_{k,c}$. Damage persists after renormalisation so it cannot be attributed to activation mismatch alone.}
\label{tab:renorm_ablation}
\begin{tabular}{lcccc}
\toprule
\Th{Method}      & $L_{0}$ & $L_{1}$ & $L_{20}$ & $L_{23}$ \\
\midrule
\Th{Reinit}      & $+167.4$ & $+60.7$ & $+33.8$ & $+48.5$ \\
+ \Th{Renorm}   & $+101.0$ & $+63.1$ & $+14.3$ & $+19.0$ \\
\bottomrule
\end{tabular}
\end{table}

\newpage

\section{Datasets}
\subsection{RadImageNet}
We use the  \Th{RadImageNet} as our primary diverse and large 2D radiology dataset. We were given access to its CT and MR sub-collections. Our subset covers five anatomical regions: CT abdomen,
CT thorax (lung), MR hip, MR knee, and MR shoulder. Each region is
further subdivided into per-pathology sub-folders, for a total of
$71$ pathology classes, with $67$ of them having more than 100 samples.

The provided set consists of $530,534$ 2D slice PNGs.
We randomly hold out $15{,}000$ images as validation and train on the remaining
$515,534$. Preprocessing involves center-cropping and bicubic upsampling to $256{\times}256$. The class-conditional generator is conditioned on the five
$(\text{modality}, \text{body part})$ region labels. The finer
pathology taxonomy enters only at evaluation, where the linear-probe analysis restricts to the $19$ pathology classes by filtering out rare (long-tail) pathology classes with fewer than $100$ samples in our validation set.

\begin{table}[!htb]
\centering
\caption{\textbf{The $19$ pathology classes used for the linear-probe
analysis.} Names follow the original \Th{RadImageNet} folder
labels verbatim. \Th{MR hip} contributes no class above the threshold.}
\label{tab:radin-probe-19}
\small
\setlength{\tabcolsep}{6pt}
\begin{tabular}{l c}
\toprule
\Th{Region} & \Th{Class} \\
\midrule
\multirow{5}{*}{\Th{CT abd}}
 & \ttt{renal\_lesion}   \\
 & \ttt{liver\_lesion}           \\
 & \ttt{bowel\_abnormality}      \\
 & \ttt{bowel\_inflammation}     \\
 & \ttt{post\_op}                \\
\midrule
\multirow{3}{*}{\Th{CT lung}}
 & \ttt{Airspace\_opacity}       \\
 & \ttt{Nodule}                  \\
 & \ttt{interstitial\_lung\_disease}   \\
\midrule
\multirow{7}{*}{\Th{MR knee}}
 & \ttt{chondral\_abnormality}   \\
 & \ttt{meniscal\_abnormality}   \\
 & \ttt{soft\_tissue\_fluid\_collection}  \\
 & \ttt{bone\_inflammation}      \\
 & \ttt{acl\_pathology}          \\
 & \ttt{mcl\_pathology}          \\
 & \ttt{intra}                   \\
\midrule
\multirow{4}{*}{\Th{MR shoulder}}
 & \ttt{labral\_pathology}       \\
 & \ttt{soft\_tissue\_fluid}     \\
 & \ttt{soft\_tissue\_edema}     \\
 & \ttt{supraspinatus\_pathology}  \\
\bottomrule
\end{tabular}
\end{table}

\paragraph{\Th{BraTS}.}
We use the 2024 BraTS post-treatment adult glioma
cohort, contributed by seven academic medical centers.
Each subject is acquired with four co-registered, skull-stripped 1\,mm
isotropic multi-parametric MRI volumes covering the standard clinical
contrasts: native T1 (\ttt{T1}), post-contrast T1-weighted (\ttt{T1c}),
T2-weighted (\ttt{T2}), and T2 fluid-attenuated inversion recovery
(\ttt{T2FLAIR}). \Th{BraTS} consists of $1{,}621$ training patients ($6{,}484$ 3D mpMRI volumes). The official validation split contributes a held-out cohort of $188$ patients ($752$ volumes).

We flatten each volume to 2D by keeping the axial slices 15--154 (140 per volume) and
bicubic-upsampling from $192{\times}192$ to $256{\times}256$. Every multi-channel $(\text{patient},\text{slice})$ pair is stored as a single
$4{\times}256{\times}256$ tensor in which the four co-registered
modalities occupy the channel dimension, yielding
$226,940$ training and $26,320$ validation multi-channel samples.

We use the data in two ways. In the \emph{4-channel} setup, each slice is one 4-channel image. In the \emph{single-channel} setup, each modality is a separate grayscale image (replicated to three channels for \Th{IN} initialisation), which gives $907,760$ training and $105,280$ validation single-channel images.

\section{Spatial structure metrics}
\label{app:ssm}

\begin{figure*}[!htb]
  \centering
  \includegraphics[width=\linewidth]{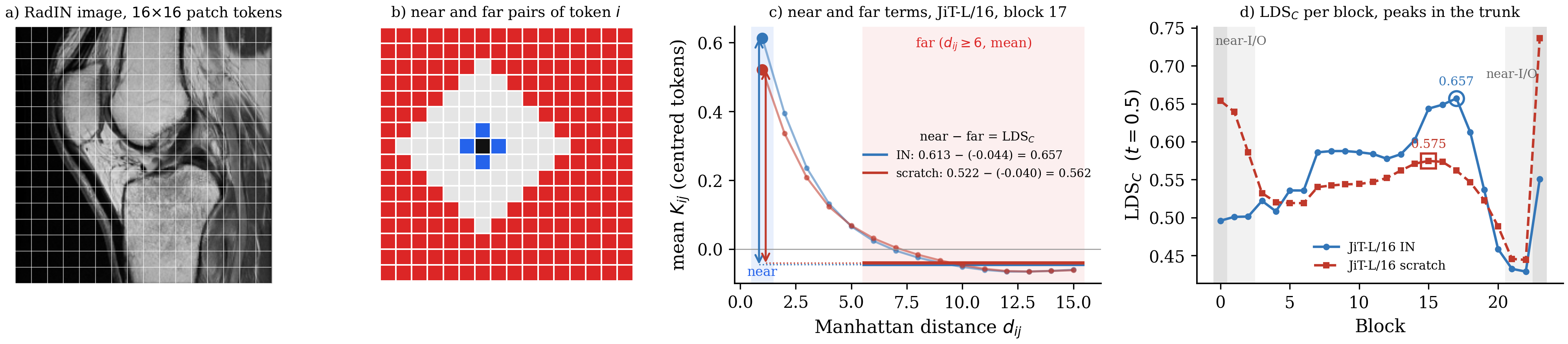}
  \caption{\textbf{LDS and \LDSc.} a) A \Th{RadIN} image and its $16{\times}16$ patch tokens.
  b) Near ($d_{ij}{<}R_{\text{n}}$, blue) and far ($d_{ij}{\geq}R_{\text{f}}$, red) pairs of token $i$.
  c) Mean cosine of mean-centred tokens vs.\ Manhattan distance for \ttt{JiT-L/16} (block $17$,
  $t{=}0.5$, $2$k \Th{RadIN} images); \LDSc\ is the near term ($d_{ij}{=}1$) minus the mean far term
  ($d_{ij}{\geq}6$). d) \LDSc\ per block. Values in the first and last blocks (dark grey) and the
  adjacent near-I/O blocks (light grey) stay close to the pixel space; excluding them places the peak
  of both models in the trunk (\autoref{app:io_exclusion}). The main paper excludes only the first
  and last block.}
  \label{fig:lds_definition}
\end{figure*}

\subsection{LDS and \LDSc: implementation}
\label{app:ldsc_impl}
We follow the definitions of \autoref{sec:ssms} and use the reference implementation of
\cite{singh2026what}. \ttt{JiT} tokenises a $256{\times}256$ image into a $16{\times}16$ grid of
$N{=}256$ patch tokens. In-context (class) tokens are removed before computing the metric.
Near pairs are adjacent tokens ($d_{ij}{=}1$, \ie, $R_{\text{n}}{=}2$; $4$ per interior token) and far pairs have
$d_{ij}{\geq}R_{\text{f}}{=}6$ ($195$ per central token), following~\cite{singh2026what}
(\autoref{fig:lds_definition} a,b). A high \LDSc means that nearby patches, \eg, the fur of a bird or a tissue in an MRI scan, have similar representations, while distant patches differ. For \LDSc, we mean-centre the tokens of each image before computing the cosine kernel, so that
\LDSc\ is the near term minus the far term of the centred tokens (\autoref{fig:lds_definition}c).
We compute \LDSc\ for every block at $t{=}0.5$ and report the peak over
blocks, excluding the I/O blocks (\autoref{fig:lds_definition} d).

\subsection{Robustness to the I/O exclusion}
\label{app:io_exclusion}
\begin{figure*}[!htb]
  \centering
  \includegraphics[width=0.9\linewidth]{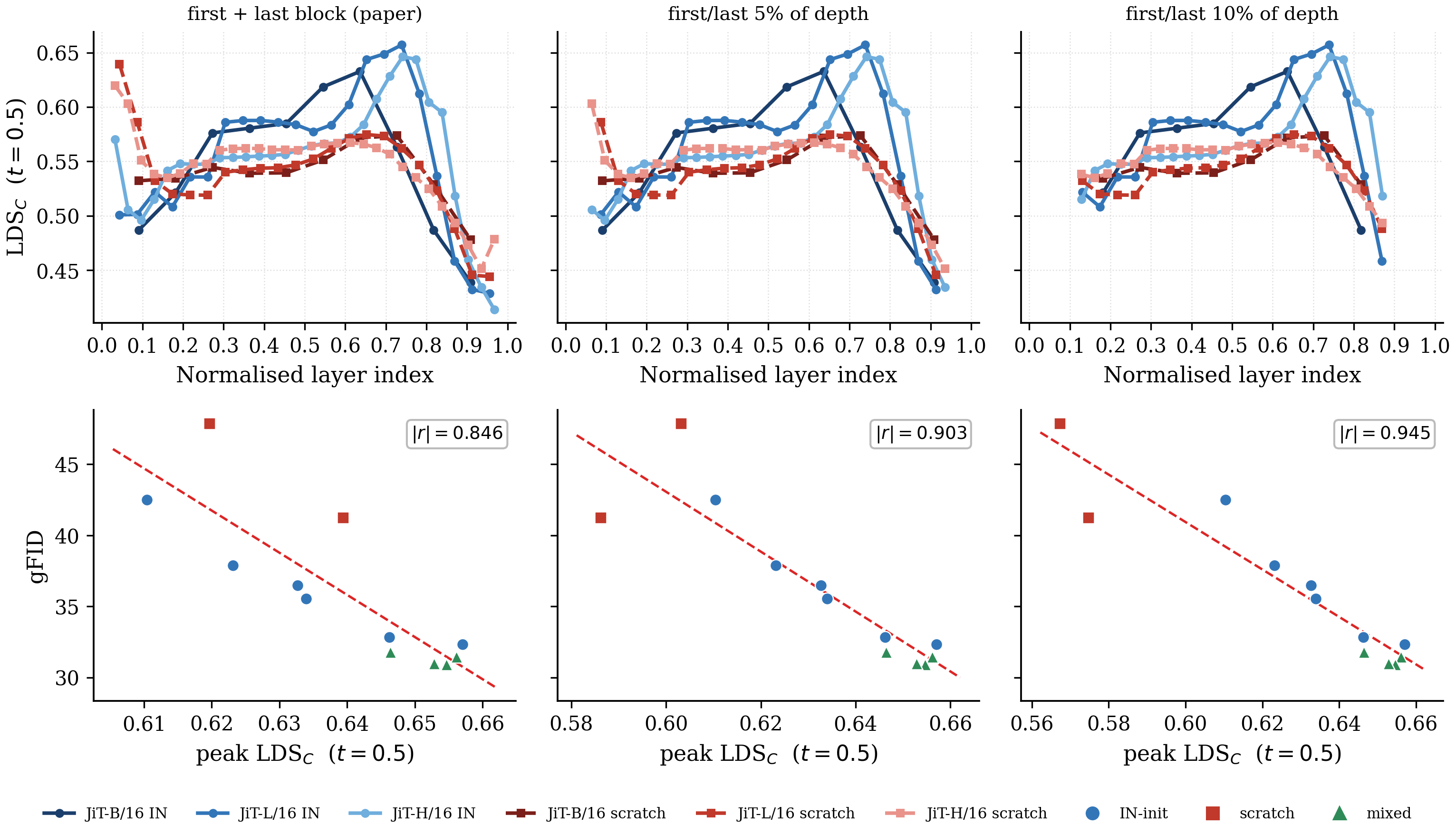}
  \caption{\textbf{Robustness to the I/O exclusion.} Top: \LDSc\ per block ($t{=}0.5$) for
  \ttt{JiT-B/L/H}, \Th{IN} (solid) and scratch (dashed), after excluding the first and last block
  (main paper), or the first/last $5\%$ or $10\%$ of depth. Bottom: peak \LDSc\ vs.\ g\Th{FID} for the
  $12$ models of \autoref{fig:lds_gfid_scatter}, with linear fit and Pearson $|r|$.}
  \label{fig:io_exclusion}
\end{figure*}

The main paper excludes only the first and last block. However, in deeper models the near-input
region spans more than one block. For example, the \ttt{L/16} and \ttt{H/16} scratch denoisers reach
their highest \LDSc\ in block $1$, directly after the input, where representations remain close to
the pixel space (\autoref{fig:lds_definition}d, \autoref{fig:io_exclusion} top). This reduces their
\Th{IN} vs.\ scratch gap ($2.8\%$ and $4.3\%$ for \ttt{L} and \ttt{H}, vs.\ $10.3\%$ for \ttt{B}).
An alternative is to exclude a fixed fraction of depth, \eg, $5\%$ or $10\%$ at each end, \ie, removing these blocks and
focusing on the mid-layers.
The scratch peaks shift into the trunk (\ttt{L/16} block $15$,
\ttt{H/16} block $19$ at $10\%$), the gap becomes similar across sizes ($10.3/14.3/13.9\%$ at
$10\%$), and the correlation with g\Th{FID} increases from $|r|{=}0.846$ to $0.903$ and $0.945$
(\autoref{fig:io_exclusion} bottom). The \Th{IN} peaks do not change. Therefore, our conclusions hold under all three rules, and the first/last-block rule is the conservative choice.

\section{Evaluation Metrics}

\paragraph{Evaluation setup.}
All quantitative metrics are computed with the
\ttt{dgm-eval} suite, using the
\Th{Inception-V3} feature extractor.
Evaluation features are extracted in \ttt{fp32}.

\paragraph{\Th{FID}}
Measures the feature distance between the distributions of real and
generated images. It uses the \Th{Inception-V3} network and computes
the Fr\'echet distance under the assumption that both feature
distributions are multivariate Gaussian. Lower is better.

\paragraph{Density \& Coverage}
Refine precision / recall to be more robust to outliers and large
feature variance. \Th{Dens.}\ counts how many real-sample
$k$-NN neighbourhoods contain each generated sample (fidelity);
\Th{Cov.}\ is the fraction of real samples whose neighbourhood contains
at least one generated sample (diversity). Higher is better.

\paragraph{Spectral profile.}
Diagnostic, not a headline metric. We take the 2D-FFT of each image,
radially average the power to obtain $P(f)$ as a function of spatial
frequency, and plot the ratio $P_{\text{gen}}(f)/P_{\text{real}}(f)$:
$1$ matches the real spectrum at that frequency, $<\!1$ is
under-production, $>\!1$ is over-shoot. We only use it for
within-family comparisons. We do not compare across architectures, because their spectra differ for reasons unrelated to transfer.

\newpage

\begin{table*}[!htb]
\centering
\caption{\textbf{Latent-space \ttt{LDiT} implementation details}}
\label{tab:ldit-impl}
\small
\setlength{\tabcolsep}{6pt}
\begin{tabular}{l ccc}
\toprule
& \ttt{LDiT-XL/2} & \ttt{LDiT-XL/2} & \ttt{LDiT-XL/1} \\
\Th{Tokeniser}
 & \cellcolor{medbg}\ttt{MedVAE}
 & \cellcolor{natbg}\ttt{SDVAE}
 & \cellcolor{natbg}\ttt{VAVAE} \\
\midrule
\sectionrow{4}{\Th{Architecture}}
Tokeniser downsample & $8\times$ & $8\times$ & $16\times$ \\
Tokeniser latent channels & $4$ & $4$ & $32$ \\
Latent grid ($256{\times}256$ in) & $32{\times}32$ & $32{\times}32$ & $16{\times}16$ \\
Latent patch size & $2$ & $2$ & $1$ \\
Token count & $256$ & $256$ & $256$ \\
Transformer blocks & $28$ & $28$ & $28$ \\
Hidden dim & $1152$ & $1152$ & $1152$ \\
Attention heads & $16$ & $16$ & $16$ \\
MLP ratio & $4$ & $4$ & $4$ \\
QK-norm & \ch & \ch & \ch \\
Activation & \Th{SwiGLU} & \Th{SwiGLU} & \Th{SwiGLU} \\
Positional emb. & \Th{RoPE} & \Th{RoPE} & \Th{RoPE} \\
Normalisation & \Th{RMSNorm} & \Th{RMSNorm} & \Th{RMSNorm} \\
Trainable params & $675$M & $675$M & $675$M \\
\midrule
\sectionrow{4}{\Th{Optimisation \& training}}
Transport & rect. flow & rect. flow & rect. flow \\
Prediction parameterisation & velocity ($\mathrm{v}$) & velocity ($\mathrm{v}$) & velocity ($\mathrm{v}$) \\
Loss & MSE on $\mathrm{v}$ & MSE on $\mathrm{v}$ & MSE on $\mathrm{v}$ \\
Velocity Directional Loss (aux) & \ch & \ch & \ch \\
$t$-sampling & logit-normal $(\mu{=}0,\sigma{=}1)$ & logit-normal $(\mu{=}0,\sigma{=}1)$ & logit-normal $(\mu{=}0,\sigma{=}1)$ \\
Optimizer & AdamW & AdamW & AdamW \\
$\beta_1, \beta_2$ & $0.9,\,0.95$ & $0.9,\,0.95$ & $0.9,\,0.95$ \\
Grad-norm clip & $1.0$ & $1.0$ & $1.0$ \\
Learning rate (LR) & $2{\times}10^{-4}$ & $2{\times}10^{-4}$ & $2{\times}10^{-4}$ \\
LR schedule & constant & constant & constant \\
Warm-up & none & none & none \\
Effective batch size & $384$ & $384$ & $384$ \\
Precision & bf16 & bf16 & bf16 \\
\midrule
\sectionrow{4}{\Th{Sampling}}
Sampler & Euler ODE & Euler ODE & Euler ODE \\
Sampling steps & $250$ & $250$ & $250$ \\
CFG scale & $1.0$ & $1.0$ & $1.0$ \\
\bottomrule
\end{tabular}
\\[4pt]
\footnotesize
\colorbox{medbg}{\phantom{XX}} Medical-image tokenizer \quad
\colorbox{natbg}{\phantom{XX}} Natural-image tokenizer.
\end{table*}

\newpage

\begin{table*}[!htb]
\centering
\caption{\textbf{Pixel-space \ttt{JiT} denoiser implementation details}}
\label{tab:jit-impl}
\small
\setlength{\tabcolsep}{6pt}
\begin{tabular}{l ccc}
\toprule
& \ttt{JiT-B/16} & \ttt{JiT-L/16} & \ttt{JiT-H/16} \\
\midrule
\sectionrow{4}{\Th{Architecture}}
Patch size & $16{\times}16$ & $16{\times}16$ & $16{\times}16$ \\
Token count ($256{\times}256$ in) & $256$ & $256$ & $256$ \\
Transformer blocks & $12$ & $24$ & $32$ \\
Hidden dim & $\phantom{0}768$ & $1024$ & $1280$ \\
Attention heads & $12$ & $16$ & $16$ \\
MLP ratio & $4$ & $4$ & $4$ \\
QK-norm & \ch & \ch & \ch \\
Activation & \Th{SwiGLU} & \Th{SwiGLU} & \Th{SwiGLU} \\
Positional emb. & sin-cos $+$ \Th{RoPE} & sin-cos $+$ \Th{RoPE} & sin-cos $+$ \Th{RoPE} \\
Normalisation & \Th{RMSNorm} & \Th{RMSNorm} & \Th{RMSNorm} \\
Trainable params & $131$M & $459$M & $952$M \\
\midrule
\sectionrow{4}{\Th{Optimisation \& training}}
Transport & rect. flow & rect. flow & rect. flow \\
Prediction parameterisation & $x$-prediction & $x$-prediction & $x$-prediction \\
Loss & MSE on $\mathrm{v}$ & MSE on $\mathrm{v}$ & MSE on $\mathrm{v}$ \\
$t$-sampling & logit-normal $(\mu{=}{-}0.8,\sigma{=}0.8)$ & logit-normal $(\mu{=}{-}0.8,\sigma{=}0.8)$ & logit-normal $(\mu{=}{-}0.8,\sigma{=}0.8)$ \\
Optimizer & AdamW & AdamW & AdamW \\
$\beta_1, \beta_2$ & $0.9,\,0.95$ & $0.9,\,0.95$ & $0.9,\,0.95$ \\
Learning Rate (LR) & $5{\times}10^{-5}$ & $5{\times}10^{-5}$ & $5{\times}10^{-5}$ \\
LR schedule & constant & constant & constant \\
Warm-up & $5$ epochs & $5$ epochs & $5$ epochs \\
Effective batch size & $288$ & $288$ & $144$ \\
Precision & bf16 & bf16 & bf16 \\
\midrule
\sectionrow{4}{\Th{Sampling}}
Sampler & Heun & Heun & Heun \\
Sampling steps & $50$ & $50$ & $50$ \\
CFG scale & $1.0$ & $1.0$ & $1.0$ \\
\bottomrule
\end{tabular}
\\[4pt]
\end{table*}

\newpage

\begin{figure*}[!htb]
    \centering
    \includegraphics[width=0.99\linewidth]{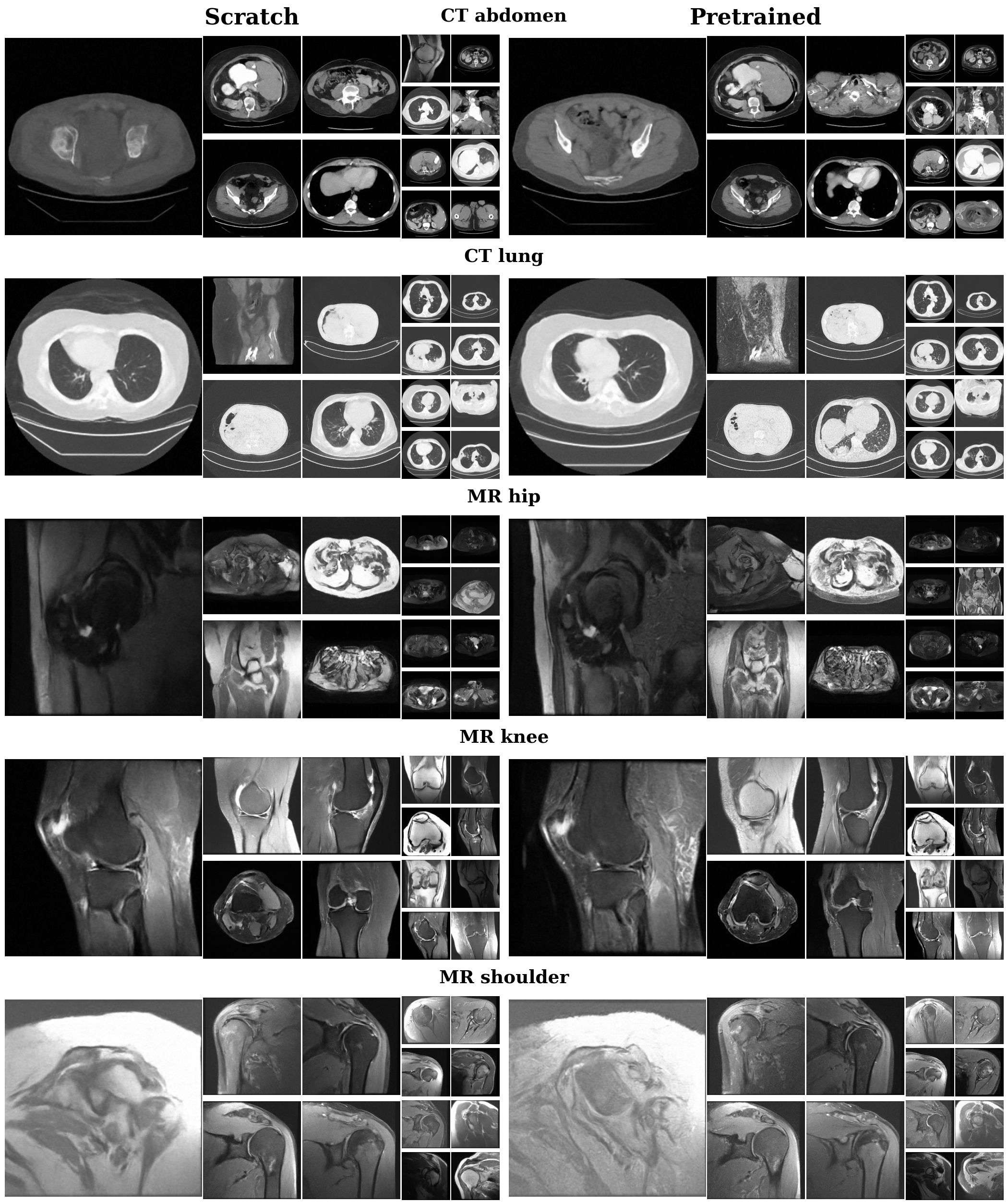}
    \caption{\textbf{\ttt{JiT-B}}. Uncurated results with cfg 1.0 on \Th{RadIN}}
\end{figure*}

\begin{figure*}[!htb]
    \centering
    \includegraphics[width=0.99\linewidth]{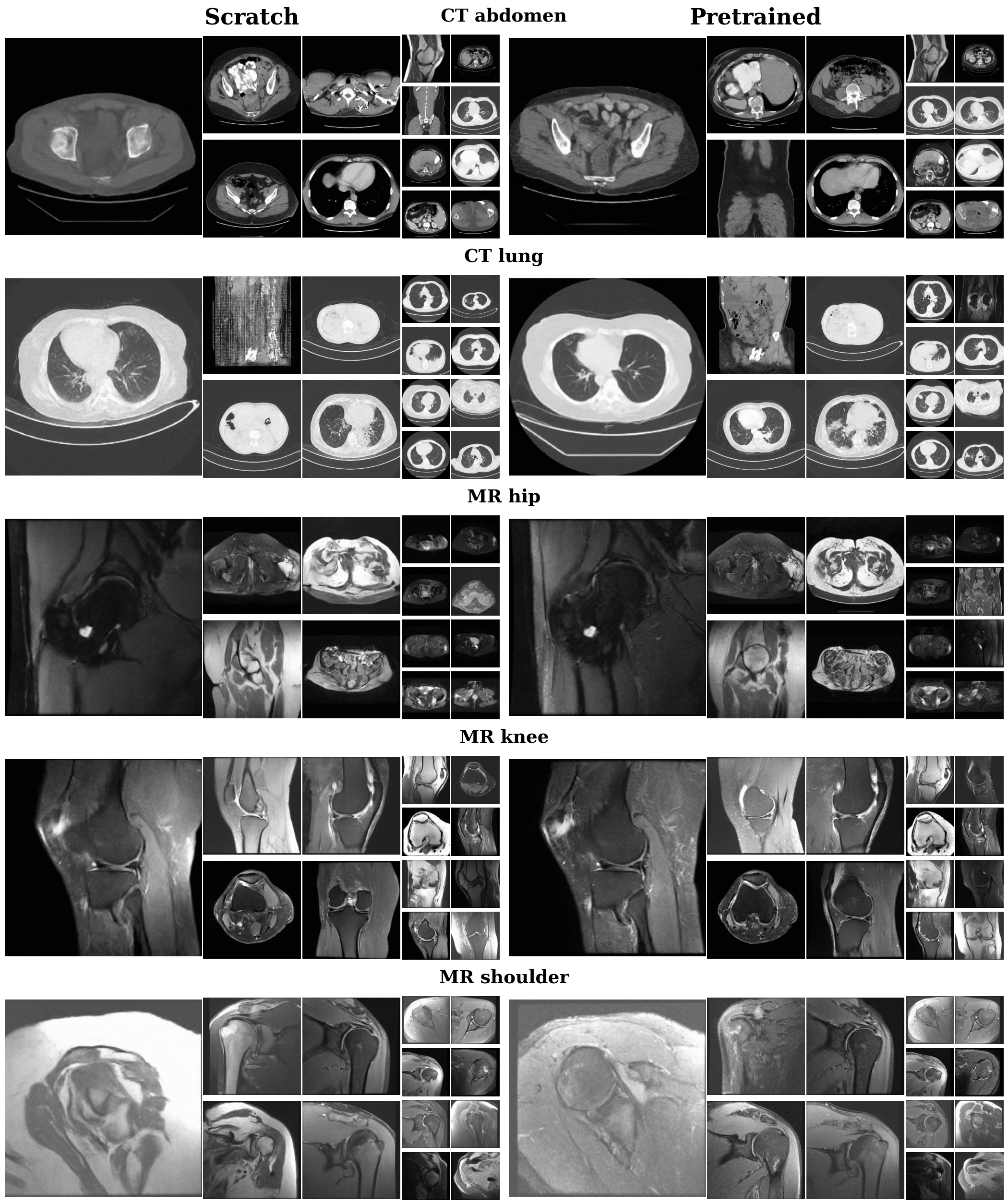}
    \caption{\textbf{\ttt{JiT-L}}. Uncurated results with cfg 1.0 on \Th{RadIN}}
\end{figure*}

\begin{figure*}[!htb]
    \centering
    \includegraphics[width=0.99\linewidth]{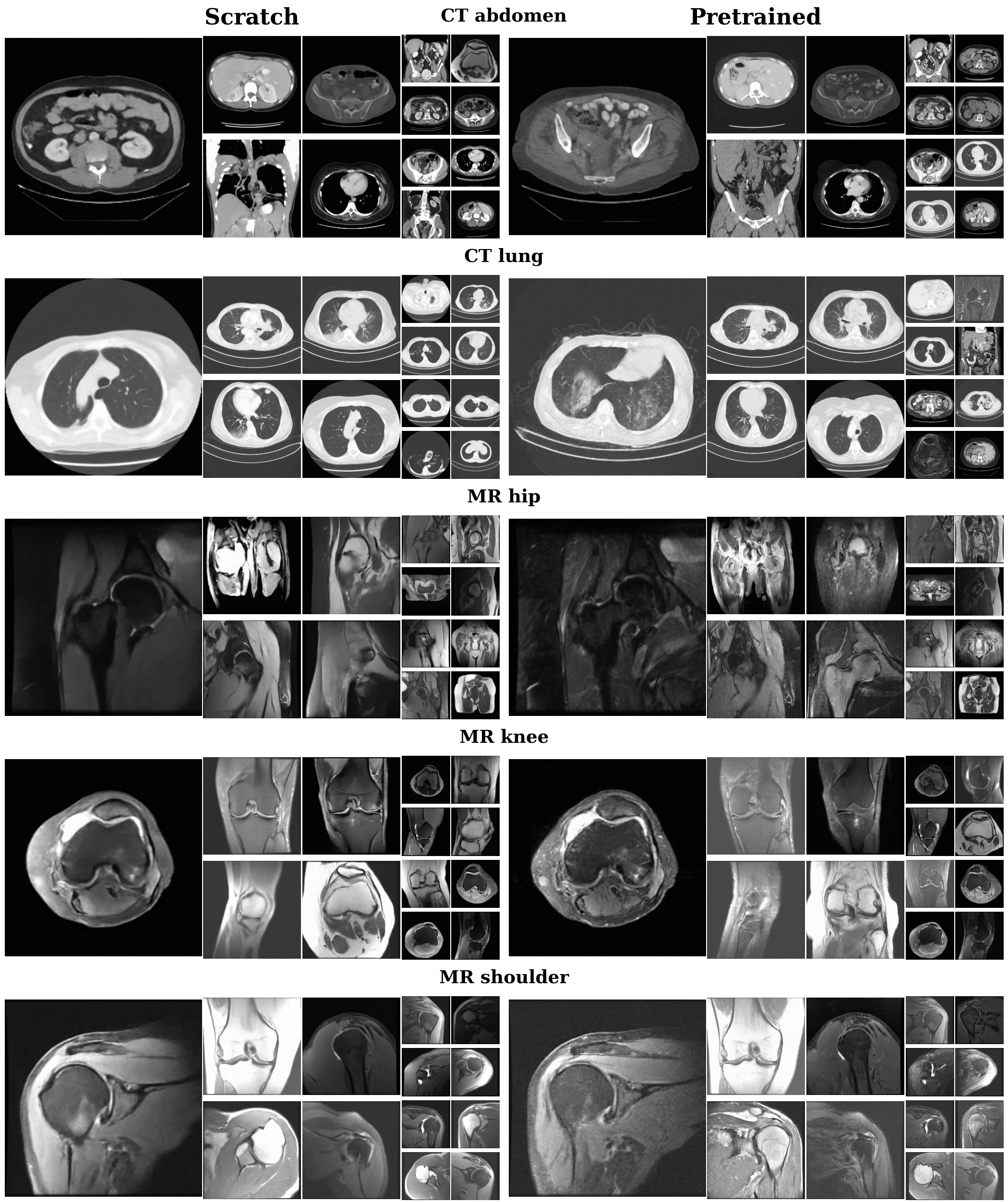}
    \caption{\textbf{\ttt{JiT-H}}. Uncurated results with cfg 1.0 on \Th{RadIN}}
\end{figure*}

\begin{figure*}[!htb]
    \centering
    \includegraphics[width=0.71\linewidth]{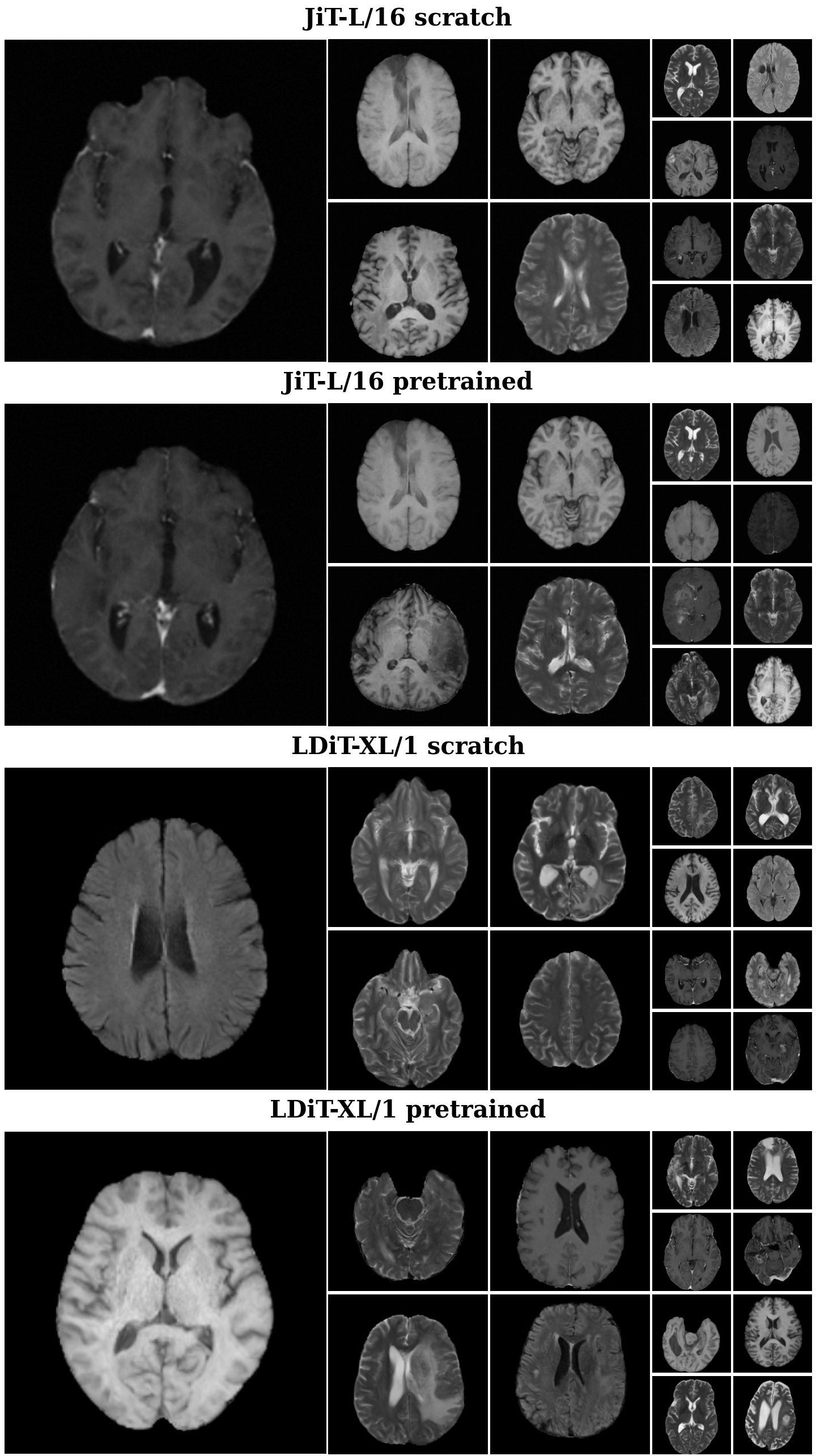}
    \caption{Uncurated results with cfg 1.0 on \Th{BraTS}}
\end{figure*}

\end{document}